%% file: arxiv.tex
\documentclass[letterpaper]{article} 

\usepackage[sort&compress,numbers]{natbib}

\usepackage[draft]{aaai2026_arxiv}  
\usepackage{times}  
\usepackage{helvet}  
\usepackage{courier}  
\usepackage[hyphens]{url}  
\usepackage{graphicx} 
\usepackage{caption} 
\usepackage{algorithm}
\usepackage{algorithmic}

\usepackage{newfloat}
\usepackage{listings}
\DeclareCaptionStyle{ruled}{labelfont=normalfont,labelsep=colon,strut=off} 
\floatstyle{ruled}
\newfloat{listing}{tb}{lst}{}
\floatname{listing}{Listing}
\title{Thinking With Videos: Multimodal Tool-Augmented Reinforcement Learning for Long Video Reasoning}
\author{
    Haoji Zhang$^{1}$\equalcontrib\quad
    Xin Gu$^{2}$\equalcontrib\quad
    Jiawen Li$^{3}$\quad 
    Chixiang Ma$^{3}$\quad 
    Sule Bai$^{1}$\quad 
    Chubin Zhang$^{1}$ \\
    Bowen Zhang$^{3}$\quad 
    Zhichao Zhou$^{3}$\quad 
    Dongliang He$^{3}$\quad 
    Yansong Tang$^{1}$\thanks{Corresponding author.}
}
\affiliations{
    {\normalsize 
    $^{1}$Tsinghua Shenzhen International Graduate School, Tsinghua University\\
    $^{2}$University of Chinese Academy of Sciences\quad
    $^{3}$Bytedance Intelligent Creation
    }
}

\usepackage{bibentry}
\usepackage{booktabs}
\usepackage{amsmath} 
\usepackage{subcaption}
\usepackage{multirow}
\usepackage{makecell}
\usepackage[dvipsnames,table]{xcolor}
\usepackage{amssymb}
\usepackage{pifont}
\definecolor{iccvblue}{rgb}{0.21,0.49,0.74}
\usepackage[pagebackref,breaklinks,colorlinks,allcolors=iccvblue]{hyperref}
\usepackage{cleveref}

\definecolor{lightblue}{RGB}{225,225,225}  
\crefname{figure}{Fig.}{Figs.}
\Crefname{figure}{Fig.}{Figs.}
\crefname{table}{Tab.}{Tabs.}
\Crefname{table}{Tab.}{Tabs.}
\crefname{equation}{Eq.}{Eqs.}
\Crefname{equation}{Eq.}{Eqs.}
\crefname{section}{Sec.}{Secs.}
\Crefname{section}{Sec.}{Secs.}
\crefname{algorithm}{Alg.}{Algs.}
\Crefname{algorithm}{Alg.}{Algs.}
\crefname{appendix}{Sec.}{Secs.}
\Crefname{appendix}{Sec.}{Secs.}

\begin{document}

\maketitle

\begin{abstract}
The video reasoning ability of multimodal large language models (MLLMs) is crucial for downstream tasks like video question answering and temporal grounding.
While recent approaches have explored text-based chain-of-thought (CoT) reasoning for MLLMs, these methods often suffer from limited cross-modal interaction and increased hallucination, especially with longer videos or reasoning chains.
To address these challenges, we propose \textbf{V}ideo \textbf{I}ntelligence via \textbf{T}ool-\textbf{A}ugmented \textbf{L}earning (\textbf{VITAL}), a novel end-to-end agentic video reasoning framework. 
With a visual toolbox, the model can densely sample new video frames on demand and generate multimodal CoT for precise long video reasoning.
We observe that temporal grounding and question answering are mutually beneficial for video understanding tasks.
Therefore, we construct two high-quality multi-task video reasoning datasets MTVR-CoT-72k for supervised fine-tuning and MTVR-RL-110k for reinforcement learning.
Moreover, we propose a Difficulty-aware Group Relative Policy Optimization algorithm (DGRPO) to mitigate difficulty imbalance in multi-task reinforcement learning.
Extensive experiments on 11 challenging video understanding benchmarks demonstrate the advanced reasoning ability of VITAL, outperforming existing methods in video question answering and temporal grounding tasks, especially in long video scenarios.
Code is available at \url{https://zhang9302002.github.io/thinkingwithvideos-page/}.
\end{abstract}

\input{main_content_arxiv}

\clearpage
{
    \small
    \bibliography{aaai2026}
}


\clearpage
\appendix
\input{supp_content}

\end{document}

%% file: main_content_arxiv.tex
\section{Introduction}
Video understanding is a fundamental challenge in artificial intelligence, with wide-ranging applications such as recommendation \cite{huang2016rec2,lee2017rec3}, 
smart surveillance \cite{tsakanikas2018sur1,chen2019sur2}, 
generation \cite{liu2024planposture, dai2024motionlcm},
visual navigation \cite{han2025roomtour3d, lin2025evolvenav},
and autonomous driving \cite{yang2024auto1,wang2024auto2}. 
Recent advances in multimodal large language models (MLLMs) \cite{bai2025qwen25vl, li2024llavaov, team2025kimivl, hurst2024gpt4o, comanici2025gemini25} have significantly improved the ability to jointly process visual and textual information, opening new opportunities for complex video reasoning tasks.

Video reasoning refers to inferring objects, relationships, events and causality from video content, often requiring multi-step and temporal understanding \cite{zhou2018sthsth, CLEVRER2020ICLR}. 
This capability is fundamental for downstream tasks like video question answering \cite{fu2025videomme}, temporal grounding \cite{gao2017charades}, spatial-temporal grounding \cite{gu2024contextstvg} and captioning \cite{shen2023accurate, chai2024auroracap}, where accurate reasoning over dynamic scenes enables more precise and informative results.
Inspired by the success of DeepSeek-R1 \cite{guo2025deepseekr1} in enhancing the reasoning ability of LLMs through reinforcement learning (RL), some works apply Group Relative Policy Optimization (GRPO) post-training to improve MLLMs' reasoning ability on images \cite{liu2025visualrft, shen2025vlmr1, bai2025univgr1}
and videos \cite{feng2025videor1, li2025videochatr1}.

\begin{figure}[t]
    \centering
    \includegraphics[width=1\linewidth]{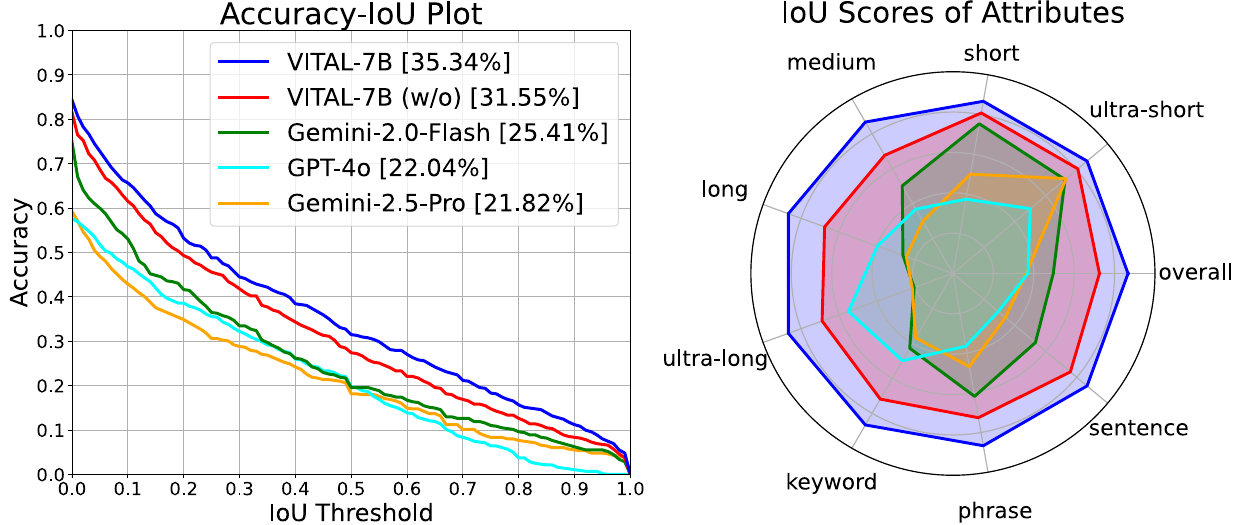}
    \caption{Performance on long video temporal grounding benchmark VUE-TR. 
    VITAL-7B (w/o) denotes VITAL-7B without toolbox.
    VITAL-7B achieves state-of-the-art.
    }
    \label{fig:teaser}
\end{figure}

\begin{figure*}[t]
    \centering
    \includegraphics[width=1\linewidth]{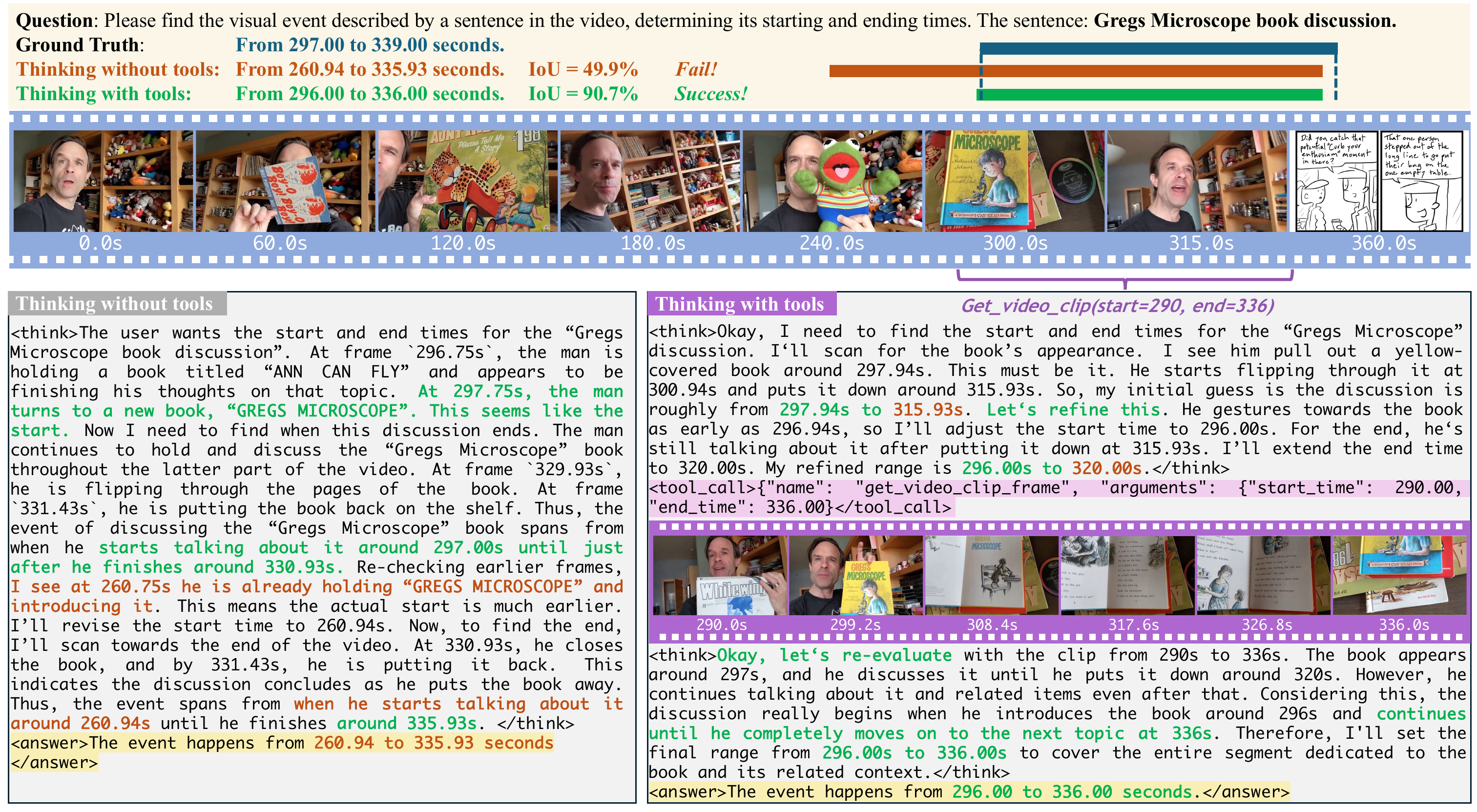}
    \caption{Comparison between text-based CoT (left) and multimodal CoT (right) on temporal grounding task.
    Green text denotes correct inference and orange text denotes wrong inference.
    ``Thinking with tools'' reduces hallucination in the reasoning process by integrating relevant, densely sampled video clip frames into multimodal CoT, resulting in more accurate grounding.
    }
    \label{fig:case}
\end{figure*}

Despite these advances, most existing MLLMs only rely on \textbf{text-based chain-of-thought} (CoT) reasoning for video understanding \cite{fei2024videoofthought, shi2025aotd, feng2025videor1, li2025videochatr1}. 
While effective in some scenarios, such approaches often suffer from two major limitations: 
(1) insufficient cross-modal interaction, which restricts the model's ability to fully leverage visual information during reasoning, and 
(2) increased hallucination, especially when handling long videos or long reasoning chains.
Thus, a key question arises: \textit{How can we enable MLLMs to perform effective reasoning over long videos, with strong cross-modal interaction and minimal hallucination?}

To address these challenges, we propose to evolve from text-based CoT to \textbf{multimodal CoT} reasoning, where the model can explicitly incorporate visual tools and dynamically attend to relevant video content throughout the reasoning process (\Cref{fig:case}). This improved paradigm enables more accurate and interpretable video understanding, 
as shown in \Cref{fig:teaser},
which highlight the effectiveness of our approach and confirm that integrating multimodal CoT is a key factor in advancing long video understanding.

Motivated by this insight, we propose \textbf{V}ideo \textbf{I}ntelligence via \textbf{T}ool-\textbf{A}ugmented \textbf{L}earning (\textbf{VITAL}), a novel end-to-end agentic framework for video multimodal (CoT) reasoning. It enables efficient multi-round multimodal tool-augmented training and evaluation. As shown in \Cref{fig:method}, VITAL consists of a MLLM and a visual toolbox, which allows the model to actively sample new frames and extract relevant multimodal information on demand during reasoning. This enables the model to focus its attention on critical temporal segments, effectively bridging the gap between textual reasoning and visual evidence and thus reducing hallucination.
To enhance the model's  multimodal reasoning abilities for efficient tool calling, we first conduct a cold-start phase, followed by reinforcement learning. In both phases, the model is jointly optimized on temporal grounding, video question answering, and grounded question answering tasks.
To support this training paradigm, we construct two large-scale, high-quality training datasets tailored for multi-task video reasoning: \textbf{MTVR-CoT-72k} for supervised fine-tuning (SFT) and \textbf{MTVR-RL-110k} for reinforcement learning (RL), as shown in \Cref{fig:dataset}.
Moreover, we propose a novel Difficulty-aware Group Relative Policy Optimization (\textbf{DGRPO}) algorithm to mitigate the difficulty imbalance in multi-task reinforcement learning. By adjusting the reward scale based on task difficulty and sample difficulty, DGRPO ensures adaptive difficulty balancing, 
leading to more stable training and improved generalization ability.
Our contributions include:
\begin{itemize}
    \item We design a tool-augmented learning framework that allows MLLMs to sample frames on demand with a visual toolbox and generate multimodal CoTs, which enables efficient training and evaluation.
    \item We construct two large-scale, high-quality multi-task video reasoning datasets: MTVR-CoT-72k and MTVR-RL-110k for comprehensive video reasoning learning.
    \item We introduce a Difficulty-aware Group Relative Policy Optimization algorithm to address difficulty imbalance in multi-task reinforcement learning.
    \item Extensive experiments on 11 benchmarks and ablation studies demonstrate that VITAL achieves superior performance in long video understanding, video question answering and temporal grounding tasks.
\end{itemize}

\section{Related Works}

\subsection{Reasoning-Enhanced MLLMs}

\begin{figure*}[t]
    \centering
    \includegraphics[width=1\linewidth]{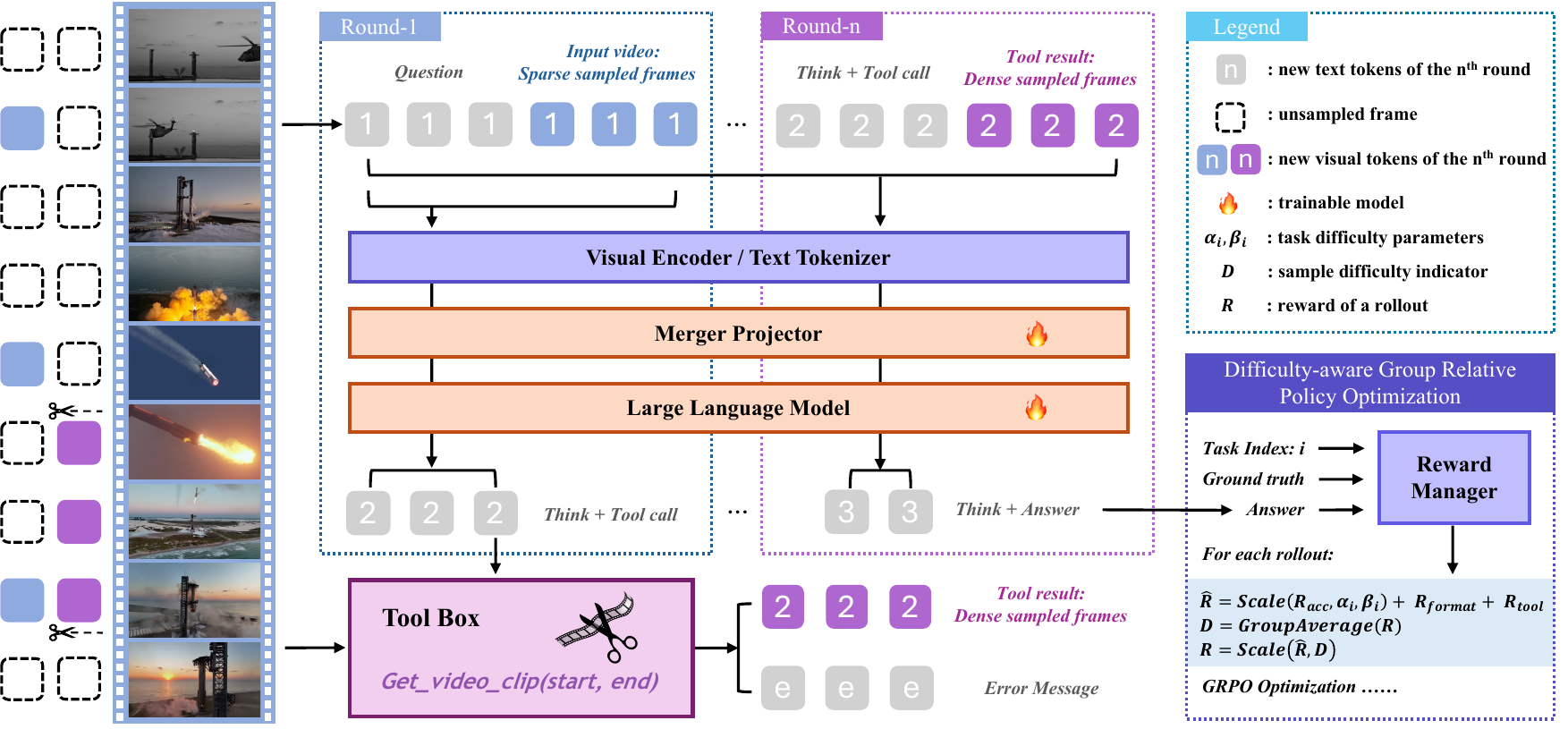}
    \caption{Overview of the Video Intelligence Tool-Augmented Learning (VITAL) framework.
    In the multi-round generation process,
    the model can attend to video tools adaptively and integrate the tool result to form a multimodal CoT. The model is optimized with Difficulty-aware Group Relative Policy Optimization (DGRPO).
    }
    \label{fig:method}
\end{figure*}

Recently, OpenAI-o1 \cite{jaech2024openaio1} and DeepSeek-R1 \cite{guo2025deepseekr1} demonstrate that RL-based post-training can further enhance the reasoning abilities of LLMs.
Following the rule-based outcome reward modeling and Group Relative Policy Optimization (GRPO) of DeepSeek-R1, some works apply similar post-training paradigms to MLLMs to enhance multimodal reasoning ability in many tasks like:
mathematical and scientific image VQA \cite{peng2025lmmr1, huang2025visionr1}; 
image segmentation and grounding \cite{liu2025seg, bai2025univgr1, shen2025vlmr1, liu2025visualrft, wang2025vgr, wang2025traceable, yang2024lavt};
video reasoning VQA \cite{feng2025videor1, wang2025videorft, li2025temporalrlt, cheng2025videoholmes};
video spatial or temporal grounding \cite{wang2025timer1, li2025videochatr1, park2025deepvideor1, ge2025hunyuanvideo7b}.
\textbf{Different from} prior works that rely on text-based CoT reasoning, we leverage multimodal CoT reasoning to effectively improve the video understanding ability of MLLMs.

\subsection{Tool-Augmented LLMs}

Recent advances in large language models (LLMs) \cite{kimiteam2025kimik2, yang2025qwen3} have shown that equipping models with external tools can enhance their capabilities beyond pure text understanding and generation, and learn to interact with the world. 
For example, using visual tools can introduce vision ability to LLMs \cite{wu2023visualchatgpt, yang2023gpt4tools}. 
Based on DeepSeek-R1 \cite{guo2025deepseekr1} style post-training, many works explore using code execution tools for mathematical and coding tasks \cite{wang2024executable, li2025start, li2025torl, feng2025retool}.
Some works leverage search engines as tools for deep search \cite{jin2025searchr1} and deep research \cite{zheng2025deepresearcher}.
In the multimodal domain, several works apply visual foundation models \cite{liu2024llavaplus}, image editing tools \cite{hu2024visualsketchpad, fu2025refocus}, or spatial grounding tools \cite{wu2024visualsearch, li2025dyfo, wei2025perceptioninreflection} to enhance visual reasoning ability.
FAST \cite{sun2025fast} and MVoT \cite{li2025MVoT} incorporate visual evidence in the reasoning process to form a multimodal CoT for image tasks. 
Inspired by DeepSeek-R1 \cite{guo2025deepseekr1} and OpenAI-o3 \cite{openai2025o3}, recent works DeepEyes \cite{zheng2025deepeyes} and OpenThinkImg \cite{su2025openthinkimg} explore ``thinking with images'' capability by integrating image zoom-in, sketching, detection and segmentation tools to MLLMs for image reasoning tasks.
\textbf{Different from} existing methods, we focus on enhancing ``thinking with videos'' capability for long video reasoning by leveraging video grounding and clipping tools.

\subsection{Long Video Understanding}

Understanding long videos poses significant challenges for MLLMs due to the high computational complexity. 
Earlier works focus on object-centric feature extraction \cite{wu2021towards, zhang2023logo}.
MIST \cite{gao2023mist} and SEVILA \cite{yu2023sevila} introduce iterative temporal selection and answering models for long video VQA tasks.
Some MLLM works \cite{weng2024longvlm, song2024moviechat, ma2024vista, he2024vlab}, LLaMA-VID \cite{li2024llama} and Flash-VStream \cite{zhang2025flashvstream} propose MLLM visual token compression techniques for long videos.
LongVA \cite{zhang2024longva} and LongVILA \cite{chenlongvila} propose long context extension finetuning, which enables training with thousands of frames per video. However, these approaches tend to be computationally intensive.
Recent works \cite{nie2024slowfocus, hannan2025revisionllm, wang2024uni} explore the course-to-fine dynamic frame sampling strategy for efficient long video understanding.
\textbf{Different from} previous works, we propose a multimodal tool-augmented RL framework for efficient and accurate long video reasoning.

\begin{figure*}[t]
    \centering
    \includegraphics[width=1\linewidth]{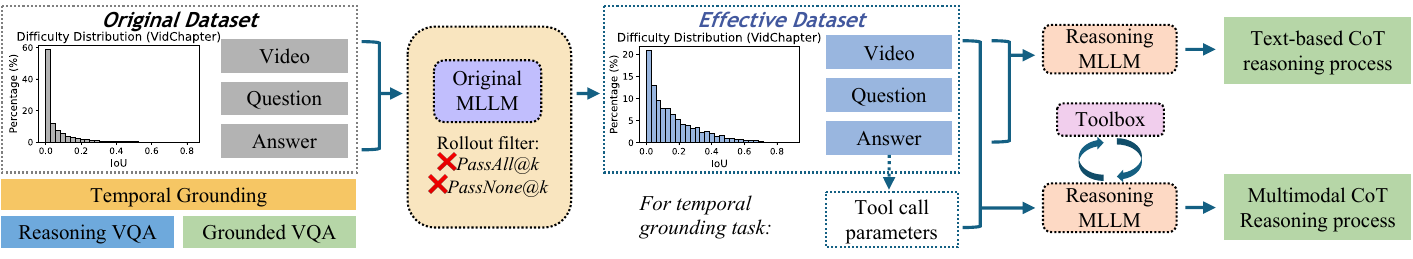}
    \caption{
    Data generation pipeline of MTVR training dataset.
    A rollout filter is applied to improve data quality.
    }
    \label{fig:dataset_pipe}
\end{figure*}

\section{Method}
\label{sec:method}
\textbf{Overview.} We propose VITAL, an end-to-end agentic video reasoning framework that enables ``thinking with videos'' by generating multimodal chain-of-thought (CoT) using a visual toolbox. 
As depicted in \Cref{fig:method}, VITAL follows the Visual Encoder-LLM architecture commonly used in MLLMs, and integrates a tool-augmented multi-round learning process with the Difficulty-aware GRPO algorithm. 
In the following subsections, we first introduce the tool-augmented learning framework of VITAL (\Cref{sec:method_tal}), then discuss the preparation of the multi-task video reasoning training data (\Cref{sec:method_mtvr}), and finally present the Difficulty-aware GRPO algorithm (\Cref{sec:method_dgrpo}), which is employed to optimize the proposed framework with the constructed training data.

\subsection{Tool-Augmented Learning Framework}
\label{sec:method_tal}

The tool-augmented learning framework operates through a multi-round generation process. In each round, the MLLM determines whether to invoke a tool from the visual toolbox.

\noindent \textbf{Multi-round Generation.}
Given a user question $\mathcal{T}_0$ and a video $\mathcal{V}_0$ as input, the VITAL model learns to reason and dynamically decide whether to call a visual tool or directly output an answer. 
If a tool is called, the next round begins after tool execution, forming a multimodal CoT.
At round-k, the model generates:
\begin{align}
\mathcal{O}_k = f_{\text{MLLM}}\big(\{\mathcal{T}_i, \mathcal{C}_i, \mathcal{V}_i\}_{i=0}^k\big)
\end{align}
After round-k generation, the parser extracts the thinking step and tool call $\mathcal{T}_{k+1}, \mathcal{C}_{k+1} = p(\mathcal{O}_k)$
or error message $\mathcal{E}_k^p$. 
Here, $\mathcal{T}_{k+1}$ is the text reasoning step and $\mathcal{C}_{k+1}$ denotes the tool call request. ($\mathcal{C}_0=\varnothing$ is empty.)
If a tool call is made, the visual toolbox executes it and returns a new video result $\mathcal{V}_{k+1}=g_{\text{tool}}(\mathcal{C}_{k+1})$ or error message $\mathcal{E}_k^g$.
$f_{\text{MLLM}}$ denotes the MLLM backbone and $g_{\text{tool}}$ denotes the visual toolbox (e.g., video clipping tools).
Specifically, when $\mathcal{C}_{k+1}==\mathcal{A}_{k+1}$, no further tool calls are made and the process terminates with the final answer.
This procedure results in a multimodal CoT trajectory 
$\tau = \left\{\mathcal{T}_1, \mathcal{C}_1, \mathcal{V}_1, \mathcal{T}_2, \mathcal{C}_2, \mathcal{V}_2, \dots, \mathcal{T}_n, \mathcal{A}_n\right\} \label{eq:tau}$.

\noindent \textbf{Visual Toolbox.}
We test several video tools in \Cref{tab:abl_tool} and identify the ``video clipping'' tool as the most effective for video temporal grounding and reasoning tasks.
As shown in \Cref{fig:method}, ``video clipping'' tool receives two time range parameters and returns a densely sampled frame sequence of the requested range.
The video clipping tool operates as:
\begin{equation}
    \mathcal{V}_{k+1} = g_{\text{clip}}(\mathcal{V}_0, t_{\text{start}}, t_{\text{end}})
\end{equation}

\subsection{Multi-Task Video Reasoning Training Data}
\label{sec:method_mtvr}
\begin{figure}[t]
    \centering
    \includegraphics[width=1\linewidth]{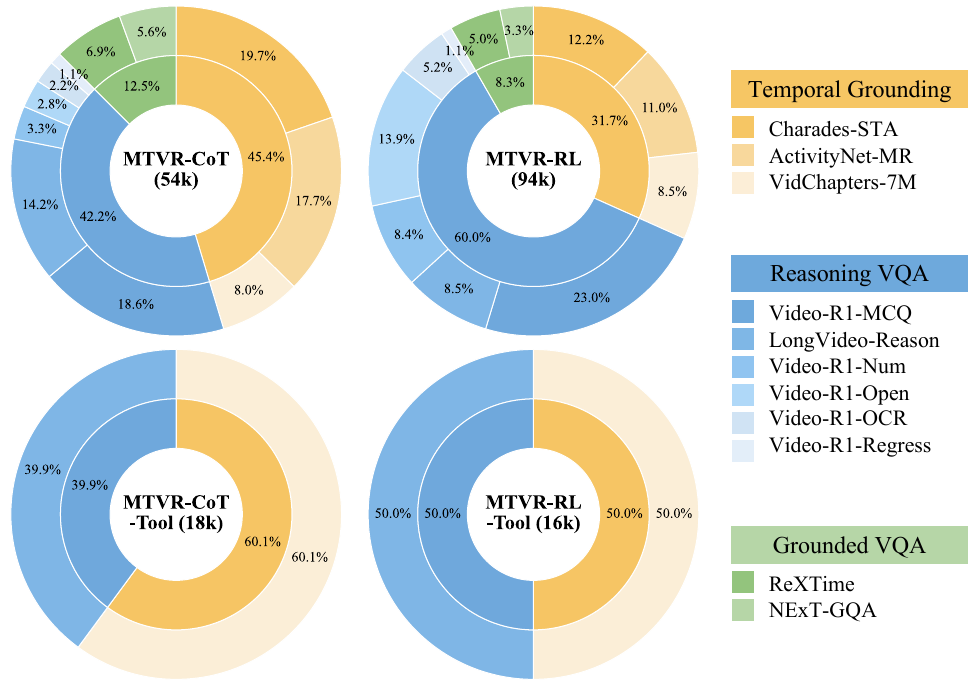}
    \caption{Task distribution of MTVR training dataset.}
    \label{fig:dataset}
\end{figure}
To enhance tool calling and multimodal reasoning, we construct two high-quality multi-task video reasoning datasets: MTVR-CoT-72k for SFT cold start and MRVR-RL-110k for RL.
These datasets cover temporal grounding, video question answering (VQA), and grounded VQA, where temporal grounding serves as the basis for question-guided video clipping, and VQA tasks evaluate general reasoning capabilities. Grounded VQA further requires the model to predict relevant time ranges and answers simultaneously.
The original data are collected from Charades-STA \cite{gao2017charades},
ActivityNet-MR \cite{krishna2017actnetcaption},
VidChapters-7M \cite{yang2023vidchapters},
Video-R1 \cite{feng2025videor1},
LongVideo-Reason \cite{chen2025longvilar1},
ReXTime \cite{chen2024rextime} and NExT-GQA \cite{xiao2024nextgqa}.

As shown in \Cref{fig:dataset_pipe}, our data generation pipeline employs a rollout filtering process to improve post-training efficiency.
For each sample, we use MLLM \cite{bai2025qwen25vl} to generate k rollouts with a high temperature to encourage diversity.
We then retain samples with moderate difficulty, filtering out those where all rollouts pass (\textit{PassAll@k}, too easy) or none pass (\textit{PassNone@k}, too hard). This results in a curated dataset with balanced difficulty.
%
After filtering out extreme data, we leverage a strong reasoning MLLM (e.g., Gemini 2.5 \cite{comanici2025gemini25}) to roll out the text-based CoT for all data and multimodal CoT for long video data.
For long video temporal grounding, we predefine visual tool parameters by adding 20\% noise to the ground truth time range. For long video QA, the reasoning MLLM autonomously selects tool parameters. We use $k=8$ and temperature $=1.0$ for both rollout generation and RL training.
Filtering criterion for different tasks and CoT generation prompts are detailed in the supplementary material.

Ultimately, we construct four data subsets for multi-stage training: \textbf{MTVR-CoT} (54k) and \textbf{MTVR-RL} (94k) for basic video reasoning, and \textbf{MTVR-CoT-Tool} (18k) and \textbf{MTVR-RL-Tool} (16k) for multi-round tool-augmented long video reasoning. 
Task and data source distributions are in \Cref{fig:dataset}.

\begin{algorithm}[t]
\caption{Difficulty-aware Reward Calculation}
\label{alg:dgrpo}
\begin{algorithmic}[1]
\REQUIRE Trajectories in a batch $\{\tau_{i,j}^k\}$, parameters $\alpha_i, \beta_i$
\FOR{each task $i$}
    \FOR{each sample $j$}
        \FOR{each rollout $k = 1, \ldots, G$}
            \STATE Compute $\mathcal{R}_{\text{acc}}(\tau_{i,j}^k)$, $\mathcal{R}_{\text{format}}(\tau_{i,j}^k)$, $\mathcal{R}_{\text{tool}}(\tau_{i,j}^k)$
            \IF{task $i$ is temporal grounding}
                \STATE $S_1 \leftarrow \mathrm{clamp}\left( \frac{\mathcal{R}_{\text{IoU}}(\tau_{i,j}^k) - \alpha_i}{\beta_i - \alpha_i}, 0, 1 \right)$
            \ELSE
                \STATE $S_1 \leftarrow \mathcal{R}_{\text{acc}}(\tau_{i,j}^k)$
            \ENDIF
            \STATE $\widehat{\mathcal{R}}(\tau_{i,j}^k) \leftarrow S_1 + \mathcal{R}_{\text{format}}(\tau_{i,j}^k) + \mathcal{R}_{\text{tool}}(\tau_{i,j}^k)$
        \ENDFOR
        \STATE $D_{i,j} \leftarrow \frac{1}{G} \sum_{k=1}^{G} \widehat{\mathcal{R}}(\tau_{i,j}^k)$
        \FOR{each rollout $k = 1, \ldots, G$}
            \STATE $w_{i,j}^k \leftarrow \mathrm{clamp}(2 - D_{i,j}, 0, 1) \times 0.5 + 0.5$
            \STATE $\mathcal{R}(\tau_{i,j}^k) \leftarrow \widehat{\mathcal{R}}(\tau_{i,j}^k) \cdot w_{i,j}^k$
        \ENDFOR
    \ENDFOR
\ENDFOR
\RETURN{$\mathcal{R}(\tau_{i,j}^k)$ for all $\tau_{i,j}^k$}
\end{algorithmic}
\end{algorithm}

\subsection{Difficulty-aware GRPO Training}
\label{sec:method_dgrpo}
To address the challenges of multi-task RL training, this section introduces our reward design and difficulty balancing strategies, 
which ensure stable optimization.

\noindent \textbf{Reward Design.}
To enable multi-task RL training, we adopt a multi-task \textit{accuracy reward} $\mathcal{R}_{\text{acc}}(\tau)$ following \cite{feng2025videor1}.
To keep stable thinking and tool calling, 
we prompt the model to format its rollout output as:
\textless think\textgreater ... \textless/think\textgreater 
\textless tool\_call\textgreater \{``name'': ..., ``arguments'': ...\}\textless/tool\_call\textgreater 
\textless think\textgreater ... \textless/think\textgreater ...
\textless answer\textgreater ... \textless/answer\textgreater 
and a strict rule-based \textit{format reward} $\mathcal{R}_{\text{format}}(\tau)$ is applied.
In order to encourage the model to attend to the toolbox for new visual information, we add another \textit{tool reward} $\mathcal{R}_{\text{tool}}(\tau)$ to a rollout if it calls at least one tool successfully.

\noindent \textbf{Difficulty Balancing.}
In our initial explorations, we supervised fine-tuned the model on MTVR-CoT and conducted GRPO training on MTVR-RL with a multi-task objective.
However, we noticed there is a phenomenon of difficulty imbalance during GRPO training.
For short video datasets and some easy tasks like multiple-choice questions, the reward increases quickly. However, for harder tasks like long video temporal grounding, the IoU reward increases more slowly, which forms \textit{task-wise difficulty imbalance}. We attribute it to the lack of discrimination for continuous IoU function.
Another observation is that as the RL training goes on, the proportion of easy samples becomes higher and that of hard samples becomes lower, as discussed in \cite{bai2025univgr1}, then the optimization soon meets a bottleneck and could not make further breakthrough, which forms \textit{sample-wise difficulty imbalance}.
Therefore, we propose a Difficulty-aware GRPO algorithm to address these problems.

\begin{table}[t]
    \centering
    \small
    \setlength{\tabcolsep}{0.7mm}
    \begin{tabular}{l|ll|ccccc}
        \Xhline{0.8pt}
         & \textbf{Train Stage} & \textbf{Style} & \makecell{\textbf{LVR}\\\textbf{Acc}} & \makecell{\textbf{VCh}\\\textbf{IoU}} & \makecell{\textbf{MMMU}\\\textbf{Acc}} & \makecell{\textbf{Cha}\\\textbf{IoU}} & \textbf{Avg} \\
        \hline
        \ding{172} & Qwen2.5-VL      & Not think    & 60.1 & 0.5  & 47.4 & 43.6 & 37.9 \\
        \ding{173} & SFT             & Not think    & 62.0 & 10.8 & 49.9 & 46.0 & 42.2 \\
        \ding{174} & SFT+GRPO        & Not think    & 63.3 & 23.5 & 50.2 & 56.2 & 48.3 \\
        \ding{175} & SFT             & Think        & 62.8 & 15.6 & 50.5 & 46.8 & 43.9 \\
        \ding{176} & SFT+GRPO        & Think        & 66.0 & 25.8 & 52.0 & 57.2 & 50.3 \\
        \ding{177} & SFT+DGRPO       & Think        & 70.2 & 28.8 & 52.1 & 57.1 & 52.1 \\
        \ding{178} & SFT+DGRPO$^*$   & Think+Tool   & 79.3 & 35.0 & 54.2 & 59.9 & 57.1 \\
        \Xhline{0.8pt}
    \end{tabular}
    \caption{Ablation study on training stages.
    Experiments show that tool-augmented DGRPO is effective for long video reasoning and temporal grounding.
    $^*$ notes this experiment repeats SFT + DGRPO for two times (four stages).
    }
    \label{tab:abl_train}
\end{table}

\begin{table*}[t]
    \centering
    \small
    \setlength{\tabcolsep}{1.5mm}
    \begin{subtable}[t]{0.33\textwidth}
        \centering
        \begin{tabular}{lccc}
            \Xhline{0.8pt}
            \multirow{2}{*}{\textbf{Model}} & \multicolumn{2}{c}{\textbf{Video-MME}} & \textbf{LVR} \\
             & \textbf{Acc} & \textbf{Long} & \textbf{Acc} \\
            \hline
            \textcolor{gray}{GPT-4o}         & \textcolor{gray}{71.9} & \textcolor{gray}{65.3}  & \textcolor{gray}{60.7} \\
            \textcolor{gray}{Gemini-1.5-Pro} & \textcolor{gray}{75.0} & \textcolor{gray}{67.4}  & \textcolor{gray}{67.3} \\
            Video-R1-7B         & 59.3 & --    & 62.7 \\
            LongVILA-R1-7B      & 62.4 & 53.3  & 67.9 \\
            Qwen2.5-VL-7B       & 62.9 & 51.0  & 60.1 \\
            \hline
            \rowcolor{lightblue}
            VITAL-7B (w/o)      & 62.5 & 51.2  & 70.2 \\
            \rowcolor{lightblue}
            \textbf{VITAL-7B}   & \textbf{64.1} & \textbf{54.0} & \textbf{79.3} \\
            \rowcolor{lightblue}
            $\Delta$Toolbox & {+1.6} & {+2.8} & {+9.1} \\
            \Xhline{0.8pt}
        \end{tabular}
        \caption{Long video question answering} 
        \label{tab:sub1}
    \end{subtable}
    \hfill
    \begin{subtable}[t]{0.36\textwidth}
        \centering
        \begin{tabular}{lccc}
            \Xhline{0.8pt}
            \multirow{2}{*}{\textbf{Model}} & \multicolumn{3}{c}{\textbf{Vid-Chapters-7M}} \\
             & $\mathbf{R@0.3}$ & $\mathbf{R@0.5}$ & $\mathbf{R@0.7}$ \\
            \hline
            VTimeLLM-7B         & 10.6 & 4.1  & 1.6 \\
            CLIP                & 10.7 & 5.2  & 2.3 \\
            M-DETR              & 37.4 & 27.3 & 17.6 \\
            ReVisionLLM         & 33.8 & 27.4 & 21.8 \\
            Qwen2.5-VL-7B       & 0.8  & 0.3  & 0.1 \\
            \hline
            \rowcolor{lightblue}
            VITAL-7B (w/o)      & 35.2 & 25.8 & 19.5 \\
            \rowcolor{lightblue}
            \textbf{VITAL-7B}   & \textbf{45.4} & \textbf{34.7} & \textbf{24.3} \\
            \rowcolor{lightblue}
            $\Delta$Toolbox        & {+10.2} & {+8.9} & {+4.8} \\
            \Xhline{0.8pt}
        \end{tabular}
        \caption{Long video temporal grounding} 
        \label{tab:sub2}
    \end{subtable}
    \hfill
    \begin{subtable}[t]{0.3\textwidth}
        \centering
        \begin{tabular}{lccc}
            \Xhline{0.8pt}
            \multirow{2}{*}{\textbf{Model}} & \multicolumn{3}{c}{\textbf{VUE-TR-Vision}} \\
             & $\mathbf{\overline{{P}}}$ & $\mathbf{\overline{{R}}}$ & $\mathbf{\overline{IoU}}$ \\
            \hline
            \textcolor{gray}{GPT-4o}              & \textcolor{gray}{41.3} & \textcolor{gray}{30.5} & \textcolor{gray}{22.0} \\
            \textcolor{gray}{Gemini-2.0-Flash}    & \textcolor{gray}{48.6} & \textcolor{gray}{38.9} & \textcolor{gray}{25.4} \\
            \textcolor{gray}{Gemini-2.5-Pro}      & \textcolor{gray}{48.2} & \textcolor{gray}{37.9} & \textcolor{gray}{21.8} \\
            Qwen2.5-VL-7B           & 44.4          & 16.0          & 12.6  \\
            \hline
            \rowcolor{lightblue}
            VITAL-7B (w/o)          & 43.7          & 60.7          & 31.6 \\
            \rowcolor{lightblue}
            \textbf{VITAL-7B}       & \textbf{47.3} & \textbf{65.0} & \textbf{35.3} \\
            \rowcolor{lightblue}
            $\Delta$Toolbox & {+3.6} & {+4.3} & {+3.7} \\
            \Xhline{0.8pt}
        \end{tabular}
        \caption{Long video temporal grounding} 
        \label{tab:sub3}
    \end{subtable}
    \caption{Performance on long video question answering  and long video temporal grounding benchmarks. Here $\mathbf{R@x}$ denotes recall at an IoU threshold of x. $\overline{\mathbf{P}}$, $\overline{\mathbf{R}}$, and $\overline{\mathbf{IoU}}$ denote Area Under Curve (AUC) values of precision, recall and Intersection over Union (IoU). 
    Gray rows denote models not open-sourced.
    VITAL-7B (w/o) denotes VITAL-7B without toolbox.
    }
    \label{tab:main}
\end{table*}

\begin{table*}[t]
    \centering
    \small
    \setlength{\tabcolsep}{1.9mm}
    \begin{tabular}{lcccccccccccc}
        \Xhline{0.8pt}
        \multirow{3}{*}{\textbf{Model}} & \multicolumn{8}{c}{\textbf{Temporal Grounding}} & \multicolumn{4}{c}{\textbf{Grounded VQA}} \\
        \cmidrule(lr){2-9} \cmidrule(lr){10-13}
        & \multicolumn{4}{c}{\textbf{Charades-STA}} & \multicolumn{4}{c}{\textbf{ActivityNet-MR}} & \multicolumn{2}{c}{\textbf{NExT-GQA}} & \multicolumn{2}{c}{\textbf{ReXTime}} \\
        & $\mathbf{R@0.3}$ & $\mathbf{R@0.5}$ & $\mathbf{R@0.7}$ & \textbf{mIoU} & $\mathbf{R@0.3}$ & $\mathbf{R@0.5}$ & $\mathbf{R@0.7}$ & \textbf{mIoU} & \textbf{mIoU} & \textbf{Acc} & \textbf{mIoU} & \textbf{Acc} \\
        \hline
        VTimeLLM-7B      & 51.0 & 27.5 & 11.4 & 31.2 & 44.0 & 27.8 & 14.3 & 30.4 & 28.8 & 17.4 & 20.1 & 36.1 \\
        TimeChat-7B      & 46.7 & 32.2 & 15.7 & 32.2 & 30.2 & 16.9 & 8.2  & 21.8 & 14.4 & 7.6  & 11.6 & 40.0 \\
        Momentor-7B      & 42.9 & 23.0 & 12.4 & 29.3 & 42.6 & 26.6 & 11.6 & 28.5 & --   & --   & --   & --   \\
        VTG-LLM-7B       & 52.0 & 33.8 & 15.7 & --   & --   & 8.3  & 3.7  & 12.0 & --   & --   & --   & --   \\
        TRACE-7B         & --   & 61.7 & 41.4 & 41.4 & 54.0 & 37.7 & 24.0 & 39.0 & --   & --   & --   & --   \\
        TimeMarker-8B    & 73.5 & 51.9 & 26.9 & 48.4 & 67.4 & 50.7 & \textbf{33.0} & 49.5 & --   & --   & --   & --   \\
        TimeZero-7B       & 78.1 & 60.8 & 35.3 & 58.1 & 58.6 & 39.0 & 21.4 & 40.5 & --   & --   & --   & --   \\
        VideoChat-R1-7B  & 82.1 & 71.7 & 50.2 & 60.8 & 51.8 & 33.4 & 17.7 & 36.6 & 32.4 & 70.6 & --   & --   \\
        TimeSearch-7B    & 73.6 & 52.4 & 24.5 & 48.6 & 61.0 & 43.0 & 26.1 & 43.9 & --   & --   & 36.7 & 76.5 \\
        Temporal-RLT-7B  & 79.6 & 67.9 & 44.1 & 57.0 & 56.9 & 38.4 & 20.2 & 39.0 & 37.3 & \textbf{78.7} & --   & --   \\
        DeepVideo-R1-7B  & --   & 71.7 & \textbf{50.6} & \textbf{61.2} & --   & 33.9 & 18.0 & 36.9 & 36.8 & 72.5 & --   & --   \\
        Qwen2.5-VL-7B    & 67.9 & 50.3 & 24.3 & 43.6 & 28.3 & 15.8 & 7.5  & 21.1 & 15.4 & 59.5 & 27.5 & 75.7 \\
        \hline
        \rowcolor{lightblue}
        VITAL-7B (w/o)   & 81.1 & 68.1 & 41.3 & 57.1 & 68.5 & 47.1 & 26.0 & 46.6 & 37.2 & 77.5 & 40.9 & 79.1 \\
        \rowcolor{lightblue}
        \textbf{VITAL-7B} & \textbf{83.1} & \textbf{72.0} & 46.7 & 59.9 & \textbf{70.9} & \textbf{50.8} & 31.6 & \textbf{49.8} & \textbf{43.0} & \textbf{78.7} & \textbf{47.6} & \textbf{80.5} \\
        \rowcolor{lightblue}
        $\Delta$Toolbox & {+2.0} & {+3.9} & {+5.4} & {+2.8} & {+2.4} & {+3.7} & {+5.6} & {+3.2} & {+5.8} & {+1.2} & {+6.7} & {+1.4} \\
        \Xhline{0.8pt}
    \end{tabular}
    \caption{Comparison of models on video temporal grounding and grounded VQA benchmarks.
    VITAL-7B (w/o) denotes VITAL-7B without toolbox.
    Best results are bold-faced.
    }
    \label{tab:temporal}
\end{table*}

Specifically, to mitigate task-wise imbalance, after getting $\mathcal{R}_{\text{acc}}(\tau)$, 
$\mathcal{R}_{\text{format}}(\tau)$ and $\mathcal{R}_{\text{tool}}(\tau)$ reward
of a rollout $\tau$,
the accuracy reward (IoU) is scaled conditioned on the difficulty of task-i when the task is temporal grounding (otherwise not scaled), as shown in \Cref{alg:dgrpo}.
Here, $\alpha_i,\beta_i$ are difficulty parameters of task $i$.
To reduce the sample-wise imbalance, we propose to calculate sample difficulty of task-i, sample-j $D_{i,j}$ by averaging the reward of all $G$ rollouts. Then we apply a soft linear scaling to $D_{i,j}$ to get the difficulty weight of sample-j (ranging from 0.5 to 1).

After this difficulty-aware reward balancing, we apply the GRPO algorithm based on rollout reward $\mathcal{R}(\tau_{i,j}^k)$ to optimize toward this objective:
\begin{align}
\mathcal{J}_{\mathrm{GRPO}}(\theta) &=
\mathbb{E}_{q \sim P(Q),\ \{\tau_k\}_{k=1}^G \sim \pi_{\theta_{\text{old}}}(\tau|q)} \notag \\
\bigg[
\frac{1}{G} \sum_{k=1}^G &
\frac{\pi_{\theta}(\tau_k|q)}{\pi_{\theta_{\text{old}}}(\tau_k|q)} A_k
- \beta\, \mathbb{D}_{\mathrm{KL}}(\pi_{\theta} \| \pi_{\mathrm{ref}})
\bigg] \\
\mathbb{D}_{KL} \left( \pi_\theta \| \pi_{ref} \right) &= 
\frac{\pi_{ref}(\tau_k|q)}{\pi_\theta(\tau_k|q)} 
- \log \frac{\pi_{ref}(\tau_k|q)}{\pi_\theta(\tau_k|q)} - 1
\end{align}
Here $q=\{\mathcal{T}_0,\mathcal{V}_0\}$ denotes the question and input video.
As presented in \Cref{tab:abl_train}, the DGRPO training is more stable than GRPO and achieves better performance on challenging long video benchmarks LongVideo-Reason (LVR), VidChapters-7M (VidCh), while reserving good accuracy on Video-MMMU and Charades-STA (Cha).
Implementation details of DGRPO are included in the supplementary.

\section{Experiments}
\subsection{Experimental Setup}
\noindent \textbf{Implementation Details.}
The VITAL-7B model is implemented with a visual encoder, a merger projector and a large language model pretrained from Qwen2.5-VL-7B \cite{bai2025qwen25vl}. 
The training framework extends the functionalities of verl \cite{zhang2024framework,sheng2025hybridflow} and vLLM \cite{kwon2023vllm}, providing additional support for multimodal tool-augmented multi-round training and evaluation.

\noindent \textbf{Training Settings.}
VITAL is trained with the AdamW optimizer \cite{loshchilov2017decoupled} and a cosine lr scheduler. The weight decay is 1e-2. The learning rate is 1e-5 for SFT and 1e-6 for RL.
The batch size is 256 for SFT and 64 for RL. The number of rollouts is 8 for DGRPO.
We train the VITAL-7B model for one epoch at each of the four stages, totaling 640 GPU hours.

\noindent \textbf{Evaluation Settings.}
We evaluated our model on long video question answering benchmark Video-MME \cite{fu2025videomme}, long video reasoning benchmark LongVideo-Reason (LVR) \cite{chen2025longvilar1}, long video temporal grounding benchmarks VidChapters-7M (VidCh) \cite{yang2023vidchapters} and VUE-TR \cite{team2025vidi}. 
We report the results on VUE-TR-Vision to test visual ability of models.

To evaluate the basic video reasoning and temporal grounding ability of our model, we further compare it with previous works on three complex video reasoning benchmarks: VSI-Bench \cite{yang2025vsibench}, Video-MMMU (MMMU) \cite{hu2025videommmu}, MMVU \cite{zhao2025mmvu};
two video temporal grounding benchmarks: Charades-STA (Cha) \cite{gao2017charades}, ActivityNet-MR \cite{krishna2017actnetcaption},
and two grounded VQA benchmarks: NExT-GQA \cite{xiao2024nextgqa}, ReXTime \cite{chen2024rextime}.
We report accuracy (Acc) for VQA tasks, mean Intersection over Union (mIoU) and recalls for temporal grounding tasks.
More evaluation details are included in the supplementary materials.

\begin{table}[t]
    \centering
    \small
    \setlength{\tabcolsep}{1.0mm}
    \begin{tabular}{lccc}
        \Xhline{0.8pt}
        \textbf{Model} & \makecell{\textbf{VSI-Bench}\\\textbf{Acc}} & \makecell{\textbf{Video-MMMU}\\\textbf{Acc}} & \makecell{\textbf{MMVU (mc)}\\\textbf{Acc}} \\
        \hline
        LongVA-7B           & 29.2 & 23.9 & --   \\
        VILA-1.5-8B         & 28.9 & 20.8 & --   \\
        LLaVA-OV-7B         & 32.4 & 33.8 & 49.2 \\
        AoTD-7B             & 28.8 & -- & -- \\
        Video-R1-7B         & 37.1 & 52.4 & 64.2 \\
        VideoRFT-7B         & 36.8 & 51.1 & 68.5 \\
        Temporal-RLT-7B     & --   & --   & 65.0 \\
        Qwen2.5-VL-7B       & 31.8 & 47.4 & 61.3 \\
        \hline
        \rowcolor{lightblue}
        VITAL-7B (w/o)      & 37.5 & 52.1 & 62.6 \\
        \rowcolor{lightblue}
        \textbf{VITAL-7B}   & \textbf{41.8} & \textbf{54.2} & \textbf{68.7} \\
        \rowcolor{lightblue}
        $\Delta$Toolbox     & {+4.3} & {+2.1} & {+6.1} \\
        \Xhline{0.8pt}
    \end{tabular}
    \caption{Comparison of models on complex video reasoning question answering benchmarks.
    VITAL-7B (w/o) denotes VITAL-7B without toolbox.
    }
    \label{tab:reason}
\end{table}

\subsection{Main Results}
\noindent \textbf{Long Video Understanding.}
As shown in \Cref{tab:main}, VITAL-7B achieves state-of-the-art performance on long video question answering and long video temporal grounding benchmarks. 
VITAL-7B outperforms previous best open-source models by a large margin on LongVideo-Reason benchmark (Acc: 79.3\% vs. 67.9\%) and VidChapters-7M (R@0.5: 34.7\% vs. 27.4\%). 
The improvement is particularly pronounced in long video scenarios, demonstrating the effectiveness of tool-augmented multimodal CoT reasoning.

\noindent \textbf{Complex Video Reasoning.}
On challenging multi-step reasoning benchmarks such as VSI-Bench, Video-MMMU, and MMVU, VITAL-7B consistently outperforms strong baselines (\Cref{tab:reason}), confirming its advanced reasoning capability across diverse video reasoning tasks like spatial reasoning and multi-discipline knowledge learning. For clearer judgement, we test models on MMVU multiple-choice split following previous work \cite{li2025temporalrlt, feng2025videor1}. 

\noindent \textbf{Video Temporal Grounding and Grounded VQA.}
As shown in \Cref{tab:temporal},
VITAL-7B excels in short video temporal grounding benchmarks (Charades-STA and ActivityNet-MR), indicating that the proposed model has strong basic temporal grounding capability.
On grounded VQA benchmarks (NExT-GQA and ReXTime), VITAL-7B also sets new state-of-the-art performance, demonstrating the multimodal CoT design facilitates the integration of accurate temporal grounding and deep video reasoning, leading to more reliable video understanding MLLMs.

\subsection{Ablation Study}

\noindent \textbf{Effectiveness of DGRPO.}
We conduct an ablation study on training stages and training styles, e.g., with or without thinking.
As shown in row 5 and row 6 of \Cref{tab:abl_train}, 
the introduction of DGRPO improves difficult long video understanding tasks by a large margin while keeping the same short video perception ability, increasing the average score from 50.3 to 52.1. 
This demonstrates that difficulty-aware reward balancing in DGRPO helps the model better handle diverse task complexities and improves overall robustness.

\noindent \textbf{Analysis of Tool-Augmented Reinforcement Learning.}
Comparing rows 6 and 7 of \Cref{tab:abl_train}, we observe that introducing tool-augmented RL improves long video perception capability,
and leads to substantial improvements across all benchmarks. 
The delta results $\Delta \text{Toolbox}$ in the last row of \Cref{tab:main,tab:reason,tab:temporal} further highlight the consistent benefits of tool integration across diverse video reasoning tasks.
\begin{table}[t]
    \centering
    \small
    \setlength{\tabcolsep}{0.8mm}
    \begin{tabular}{l|ll|ccccc}
        \Xhline{0.8pt}
         & \textbf{Training Data} & \textbf{Size} & \makecell{\textbf{LVR}\\\textbf{Acc}} & \makecell{\textbf{VidCh}\\\textbf{mIoU}} & \makecell{\textbf{MMMU}\\\textbf{Acc}} & \makecell{\textbf{Cha}\\\textbf{mIoU}} & \textbf{Avg} \\
        \hline
        \ding{172}  & None      & 0   & 60.1 & 0.5 & 47.4 & 43.6 & 37.9 \\
        & TG        & 73k & 59.5 & 24.8 & 44.5 & 52.8 & 45.4 \\
        & RQA       & 95k & 76.5 & 4.7 & 48.9 & 37.9 & 42.0 \\
        & TG+RQA    & 168k& 78.3 & 33.8 & 52.3 & 57.6 & 55.5 \\
        \ding{178}  &TG+RQA+GQA & 182k& 79.3 & 35.0 & 54.2 & 59.9 & 57.1 \\
        \Xhline{0.8pt}
    \end{tabular}
    \caption{Analysis of training data composition.
    Combining temporal grounding, reasoning VQA and grounded VQA brings the best result.
    Experiments in line 2-5 conduct the same four stages training as Exp. \ding{178}.
    }
    \label{tab:abl_data}
\end{table}
\noindent \textbf{Analysis of Training Dataset Composition.}
We categorize MTVR-CoT-72k and MTVR-RL-110k by task type, and train VITAL-7B on different task combinations, each for four stages.
As shown in \Cref{tab:abl_data}, combining temporal grounding (TG), reasoning VQA (RQA), and grounded VQA (GQA) data yields the best overall performance, with the average score improving from 37.9\% (without training data) to 57.1\%. This result highlights that multi-task training with all three data types provides strong synergy and is crucial for robust video reasoning and temporal grounding.

\noindent \textbf{Analysis of Visual Tools.}
\Cref{tab:abl_tool} compares the impact of different visual tools on temporal grounding task.
We adopt two strong reasoning MLLMs GPT and Gemini.
We observe that adding clip caption or clip QA tools does not improve results and even leads to significant drops in mIoU. 
Video clipping empirically reduces hallucination compared to other tools
We notice all these tools could not improve the performance of GPT or Gemini. 
We suggest this is due to the zero-shot setting and the models are not optimized with visual tools.
Thus, we adopt video clipping tool by default and conduct 4-stage training to enhance tool calling ability.
Tool implementation details are in supplementary.

\begin{table}[t]
    \centering
    \small
    \setlength{\tabcolsep}{1.2mm}
    \begin{tabular}{ll|cccc}
        \Xhline{0.8pt}
        \multirow{3}{*}{\textbf{Style}} & \multirow{3}{*}{\textbf{Tool}}
        & \multicolumn{2}{c}{\textbf{Vid-Chapters-7M$^*$}} 
        & \multicolumn{2}{c}{\textbf{Charades-STA$^*$}} \\
        & & \multicolumn{2}{c}{\textbf{mIoU}} & \multicolumn{2}{c}{\textbf{mIoU}} \\
         & & GPT & Gemini & GPT & Gemini \\
        \hline
        Not think & -- & 26.3 & 28.1 & 31.5 & 32.1 \\
        Think & -- & 28.4 & 29.8 & 30.1 & 33.0 \\
        Think & Clip caption & 2.0 & 8.0 & 30.8 & 27.8 \\
        Think & Clip QA & 2.0 & 10.4 & 25.5 & 28.1 \\
        Think & Video clip & 25.5 & 26.3 & 28.8 & 31.7 \\
        \Xhline{0.8pt}
    \end{tabular}
    \caption{Ablation study on different visual tools.
    We test the zero-shot temporal grounding ability of GPT-4.1 and Gemini-2.5-Pro with different visual tools.
    $^*$ denotes using a 3k subset of the benchmark to reduce computational cost.
    }
    \label{tab:abl_tool}
\end{table}

\subsection{Qualitative Analysis}
\Cref{fig:case} presents a qualitative comparison between text-based CoT and multimodal CoT on the temporal grounding task. 
The former relies solely on textual reasoning, which leads to inaccurate self-reflection and a lower IoU score due to error accumulation and hallucination. 
In contrast, the multimodal CoT leverages visual evidence during reasoning, which allows the model to more precisely determine the temporal boundaries of the target event.
More qualitative case analyses are provided in the supplementary material.

\section{Conclusion}
In this work, we introduce VITAL, a novel tool-augmented framework that empowers MLLMs with advanced long video reasoning capabilities
, which effectively mitigates hallucination and enhances cross-modal interaction. 
We further construct two high-quality multi-task video reasoning datasets and propose the DGRPO algorithm to address difficulty imbalance in multi-task reinforcement learning. 
Our results highlight the importance of tool-augmented multimodal reasoning and provide valuable insights for future research in long video understanding.

%% file: supp_content.tex
\def\thesection{\Alph{section}}

\section*{Supplementary Material}

To facilitate a deeper understanding and reproducibility of our work, this supplementary material provides additional details on the implementation of the proposed VITAL framework, including \textbf{method} details (\Cref{sec:supp_method}), \textbf{experiment} details (\Cref{sec:supp_training}), and \textbf{dataset} construction (\Cref{sec:supp_dataset}). 
We further present extended \textbf{ablation studies} (\Cref{sec:supp_ablation}) 
and additional \textbf{case analyses} (\Cref{sec:supp_case}) 
to comprehensively evaluate the effectiveness and robustness of our approach. 
Finally, we discuss the \textbf{limitations} of our current work and outline potential directions for \textbf{future work} (\Cref{sec:supp_limitations}). 
For reference and reproducibility, we also provide the experimental \textbf{code} along with detailed documentation.

\section{Method Implementation Details}
\label{sec:supp_method}
This section describes the methodology details of the proposed VITAL framework in Sec. 3.
We first illustrate the implementation details of toolbox $g_{\text{tool}}$ and then demonstrate the details of Difficulty-aware Group Relative Policy Optimization (DGRPO).
\subsection{Tool Implementation Details}
In order to enhance the video reasoning ability of MLLMs by incorporating new evidences related to the user query, we implement three visual tools in the visual toolbox:
\begin{enumerate}
    \item \textbf{Video clip captioning tool}: This tool takes the \texttt{start} and \texttt{end} timestamps of a video clip as input, and generates a descriptive \texttt{caption} for the specified segment.
    \item \textbf{Video clip QA tool}: This tool receives the \texttt{start} and \texttt{end} timestamps of a video clip, together with a natural language \texttt{question}, as input. It outputs an \texttt{answer} to the given question based on the visual content of the specified clip.
    \item \textbf{Video clipping tool}: This tool takes the \texttt{start} and \texttt{end} timestamps as input and outputs the visual content (represented as \texttt{visual tokens}) corresponding to the selected video segment.
\end{enumerate}
\Cref{tab:tool_param} summarizes the input and output formats of these visual tools.

\begin{table}[htbp]
    \centering
    \small
    \setlength{\tabcolsep}{2mm}
    \begin{tabular}{lll}
        \toprule
        \textbf{Tool Name} & \textbf{Inputs} & \textbf{Outputs} \\
        \midrule
        Clip Captioning & start, end           & caption of the clip \\
        Clip QA      & start, end, question & answer to the question \\
        Video Clipping   & start, end           & video clip visual tokens \\
        \bottomrule
    \end{tabular}
    \caption{Inputs and outputs of visual tools}
    \label{tab:tool_param}
\end{table}

\begin{figure}[!t]
    \centering
    \includegraphics[width=1\linewidth]{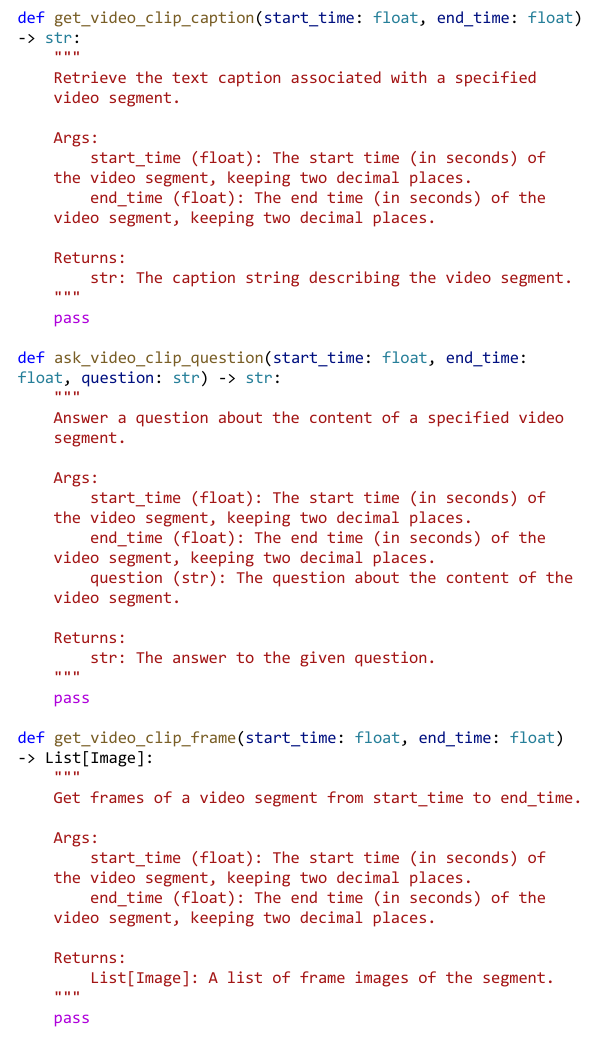}
    \caption{Python-style pseudo code for the three video reasoning tools, showing input parameters, output types, and function descriptions.}
    \label{fig:supp_tool}
\end{figure}

For all tools, the model parses and extracts tool call parameters following the Hermes-style protocol~\cite{bai2025qwen25vl, teknium2024hermes}. Each tool call is formatted as a JSON string, enclosed within the \texttt{<tool\_call>} and \texttt{</tool\_call>} tags.
After tool execution, the tool response is returned as a JSON string; if the tool outputs video tokens, the response includes a corresponding video token sequence.
In cases where tool execution fails—due to incorrect parameter formatting or other unexpected errors—an error message is returned in the form of a JSON dictionary string.

The system prompt includes tool schemas derived from pseudo code in \Cref{fig:supp_tool}.

\subsection{DGRPO Algorithm Details}
\label{sec:supp_method_dgrpo}
We propose Difficulty-aware GRPO to mitigate the task-wise difficulty imbalance and sample-wise difficulty imbalance, as discussed in Sec. 3.3 and Alg. 1.
In this subsection, we present the hyper-parameters used in our approach and provide the rationale behind their selection.

\begin{figure*}[t]
    \centering
    \includegraphics[width=0.9\linewidth]{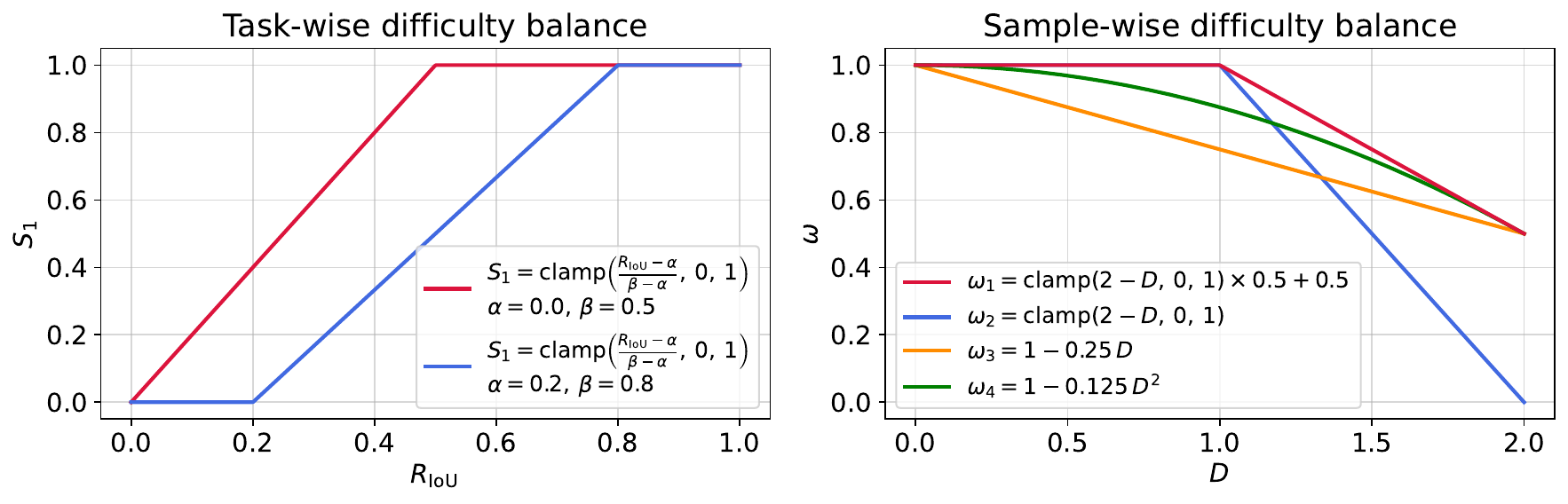}
    \caption{DGRPO clamped difficulty balance functions. Left: task-wise difficulty balance functions. Here we adopt two hyper-parameter settings for short video temporal grounding and long video temporal grounding tasks. Right: sample-wise difficulty balance functions. Here we adopt $\omega_1$ for simplicity and effectiveness.}
    \label{fig:supp_dgrpo_clamp}
\end{figure*}

\begin{table*}[t]
    \centering
    \begin{minipage}{0.56\textwidth}
        \centering
        \small
        \setlength{\tabcolsep}{1mm}
        \begin{tabular}{lllccc}
            \toprule
            \textbf{Task} & \textbf{Sub-task} & \textbf{Accuracy Reward} & \multicolumn{3}{c}{\textbf{Max Value}} \\
            \cmidrule(lr){4-6}
            & & & \textbf{Acc} & \textbf{Format} & \textbf{Tool} \\
            \midrule
            Temporal Grounding
                & --  & IoU & 1 & 0.5 & 0.5 \\
            \midrule
            \multirow{5}{*}{Reasoning VQA} 
                & MCQ       & EM        & 1 & 0.5 & 0.5 \\
                & Number    & EM        & 1 & 0.5 & 0.5 \\
                & Open-ended& Rouge     & 1 & 0.5 & 0.5 \\
                & OCR       & WER       & 1 & 0.5 & 0.5 \\
                & Regression& L1        & 1 & 0.5 & 0.5 \\
            \midrule
            \multirow{2}{*}{Grounded VQA} 
                & MCQ & (IoU + EM)/2 & 1 & 0.5 & 0.5 \\
                & Open-ended & (IoU + Rouge)/2 & 1 & 0.5 & 0.5 \\
            \bottomrule
        \end{tabular}
        \caption{Reward design of DGRPO with tools. For DGRPO without tools, tool reward is removed and the format reward ranges from 0 to 1.}
        \label{tab:supp_reward}
    \end{minipage}%
    \hfill
    \begin{minipage}{0.4\textwidth}
        \centering
        \small
        \begin{tabular}{llcc}
            \toprule
            \textbf{Task} & \textbf{Data source} & $\alpha$ & $\beta$ \\
            \midrule
            \multirow{3}{*}{Temporal Grounding}
                & Charades-STA      & 0.2 & 0.8 \\
                & ActivityNet-MR    & 0.2 & 0.8 \\
                & VidChapters-7M    & 0.0 & 0.5 \\
            \midrule
            Reasoning VQA & All     & --  & -- \\
            \midrule
            \multirow{2}{*}{Grounded VQA}
                & ReXTime           & 0.2 & 0.8 \\
                & NExT-GQA          & 0.2 & 0.8 \\
            \bottomrule
        \end{tabular}
        \caption{DGRPO hyper-parameters $\alpha$ and $\beta$ for each task and data source of MTVR-RL-110k training dataset.}
        \label{tab:supp_dgrpo_params}
    \end{minipage}
\end{table*}

\paragraph{Reward design.} 
In DGRPO reinforcement learning, we adopt three rewards: the accuracy reward, the format reward and the tool reward, as shown in \Cref{tab:supp_reward}.

For the accuracy reward, we adopt following rewards for each sub-task following \cite{feng2025videor1}:
\begin{itemize}
    \item \textbf{IoU}: the Intersection over Union value of predicted time range and ground truth time range.
    \item \textbf{EM}: Exact Match reward, is equal to 1 when the prediction matches ground truth answer exactly.
    \item \textbf{Rouge}: Recall-Oriented Understudy for Gisting Evaluation \cite{lin2004rouge}, calculated as an average of Rouge-1, Rouge-2 and Rouge-L between the prediction and the ground truth answer.
    \item \textbf{WER}: Word Error Rate, measuring the edit distance between the prediction and the ground truth answer.
    \item \textbf{L1}: calculated as $\text{clamp}(1 - |\hat y - y| / |y|,0,1)$, where, $\hat y$ is the predicted number and $y$ is the ground truth number.
\end{itemize}

The format reward is equal to 0.5 only when the model response $\tau$ matches this format exactly: 
\textless think\textgreater ... \textless/think\textgreater \ 
\textless tool\_call\textgreater ... \textless/tool\_call\textgreater \ 
\textless think\textgreater ... \textless/think\textgreater \ 
\textless answer\textgreater ... \textless/answer\textgreater 
, or 
\textless think\textgreater ... \textless/think\textgreater \ 
\textless answer\textgreater ... \textless/answer\textgreater 
If the response does not match any of these formats, the format reward is set to 0.

Similarly, the tool reward is equal to 0.5 only when the model successfully performs a tool call, otherwise it is 0.
For the second stage DGRPO without tools, the tool reward is removed and the format reward ranges from 0 to 1.

\paragraph{Task-wise difficulty balance.} 
DGRPO algorithm uses clamped linear transformations for task-wise difficulty balance.
As shown in \Cref{tab:supp_dgrpo_params} and in the left of \Cref{fig:supp_dgrpo_clamp}, we adopt two hyper-parameter settings for short video temporal grounding and long video temporal grounding tasks based on their difficulty distributions.
For multiple-choice question VQA tasks, we do nothing about task-wise difficulty balance as their reward metric is originally discrete (0 or 1).
For Gounded VQA tasks, we transform the IoU value for balancing before adding it with EM or Rouge value.

\paragraph{Sample-wise difficulty balance.} 
As presented in the right of \Cref{fig:supp_dgrpo_clamp}, we apply sample-wise difficulty balance with another clamped linear transformation $\omega_1$,
which results in a updating weight $w$ based on sample difficulty $D$.
The sample difficulty $D$ is estimated by averaging all rollout rewards $\hat{R}$ of the sample.
We adopt the piecewise linear function $\omega_1$ with the motivation to provide \textit{soft penalty to easier samples} regardless of the format reward and the tool reward. (format reward + tool reward $\leq$ 1, and they always converge to a constant value after several training steps)
We also compare other transformation functions in \Cref{sec:supp_ablation_transform}.

\begin{table*}[t]
    \centering
    \small
    \setlength{\tabcolsep}{6mm}
    \begin{tabular}{l|cccc}
        \toprule
        \textbf{Configuration} & \textbf{Stage-1} & \textbf{Stage-2} & \textbf{Stage-3} & \textbf{Stage-4} \\
        \midrule
        method & SFT & DGRPO & SFT & DGRPO \\
        using\_tools & False & False & True & True \\
        freeze\_visual\_encoder & \multicolumn{4}{c}{True} \\
        learning\_rate & 1e-5 & 1e-6 & 1e-5 & 1e-6 \\
        kl\_loss\_coef ($\beta$) & 0 & 1e-2 & 0 & 1e-2 \\
        optimizer & \multicolumn{4}{c}{AdamW} \\
        AdamW\_betas & \multicolumn{4}{c}{(0.9, 0.999)} \\
        weight\_decay & \multicolumn{4}{c}{1e-2} \\
        warmup\_ratio & 0.1 & 0 & 0.1 & 0 \\
        lr\_scheduler & \multicolumn{4}{c}{cosine} \\
        group\_size & -- & 8 & -- & 8 \\
        batch\_size & 256 & 64 & 256 & 64 \\
        mini\_batch\_size & 256 & 64 & 256 & 64 \\
        micro\_batch\_size\_per\_device & 4 & 2 & 4 & 2 \\
        number\_of\_samples & 54k & 94k & 18k & 16k \\
        number\_of\_epochs & 1 & 1 & 1 & 1 \\
        max\_num\_turns & 0 & 0 & 2 & 2 \\
        max\_total\_pixels & \multicolumn{2}{c}{64$\times$224$\times$224} & \multicolumn{2}{c}{3$\times$64$\times$224$\times$224} \\
        max\_sequence\_length & 4096 & 4096 & 10240 & 10240 \\
        max\_response\_length & 0 & 1024 & 0 & 1024 \\
        training throughput (/gpu/h) & 1408.0 & 142.9 & 1429.3 & 32.7 \\
        sample speed (s) & 2.6 & 25.2 & 2.5 & 110.3 \\
        \bottomrule
    \end{tabular}
    \caption{Training configurations.
    Here group\_size is the number of rollouts, and num\_turns is the number of tool call rounds.
    }
    \label{tab:supp_train_config}
\end{table*}

\section{Experimental Details}
\label{sec:supp_training}
In this section, we provide more details about the four-stage training procedure, the evaluation settings and metrics.
For image samples in the training dataset, we set max\_pixels = 448$\times$448. For video samples in the training dataset and evaluation datasets, we first sample them at FPS = 2 and then bound the number of frames and frame pixels according to the video length.
We adopt the number prompt technique \cite{wu2025numberit} during training and evaluation to print absolute timestamps on frames, providing additional temporal information for MLLMs for accurate temporal perception.

\subsection{Training Details}
The training configurations are listed in \Cref{tab:supp_train_config}.
Generally, we split the training procedure into two phases, post-training without tools and post-training with tools, each of which contains a cold-start SFT stage and a RL stage.
For the 3rd and the 4th stage training with tools, we only train the model on long video datasets since our motivation is to enhance long video understanding with visual tools.
We also tried training with tools from the pretrained MLLM directly, but found this paradigm is inefficient and the model is optimized slowly, since tool-augmented RL takes longer time than text-based RL because it contains multi-round generation, as shown in \Cref{tab:supp_train_config}.

In each training stage, we sample short videos with max\_frames = 64 and max\_pixels = 224$\times$224, while sampling long videos (e.g., LongVideo-Reason or Vid-Chapters-7M) with max\_frames = 256 and max\_pixels = 112$\times$112 for the initial input video. 
For densely sampled video clip, i.e., the tool results, we set max\_frames = 64 and max\_pixels = 224$\times$224 for all videos.
For DGRPO rollout generation, we set temperature = 1.0, top\_p = 1.0, group\_size = 8.

\subsection{Evaluation Details}
We evaluate the VITAL-7B model on eleven challenging benchmarks.
The evaluation details are illustrated in table \Cref{tab:supp_eval}.
In all evaluation experiments, 
we keep temperature = 0.01 and top\_p = 0.001, with max\_response\_length = 1024 and max\_num\_turns = 2, which guarantees stable and reproducible results.

\begin{table}[htbp]
    \centering
    \small
    \setlength{\tabcolsep}{0.5mm}
    \begin{tabular}{lcc}
        \toprule
        \textbf{Dataset} & \textbf{Max frames} & \textbf{Max pixels} \\
        \midrule
        Video-MME \cite{fu2025videomme}         & 1024 & 224*224 \\
        LongVideo-Reason \cite{chen2025longvilar1}  & 1024 & 224*224 \\
        Vid-Chapter-7M \cite{yang2023vidchapters}    & 1024 & 224*224 \\
        VUE-TR-Vision \cite{team2025vidi}     & 1024 & 224*224 \\
        Charades-STA \cite{gao2017charades}      & 256  & 384*384 \\
        ActivityNet-MR \cite{krishna2017actnetcaption}    & 256  & 384*384 \\
        NExT-GQA \cite{xiao2024nextgqa}          & 256  & 384*384 \\
        ReXTime \cite{chen2024rextime}           & 256  & 384*384 \\
        VSI-Bench \cite{yang2025vsibench}         & 256  & 384*384 \\
        Video-MMMU \cite{hu2025videommmu}        & 256  & 384*384 \\
        MMVU (mc) \cite{zhao2025mmvu}         & 256  & 384*384 \\
        \bottomrule
    \end{tabular}
    \caption{Evaluation configurations for each dataset.}
    \label{tab:supp_eval}
\end{table}

\section{Dataset Construction Details}
\label{sec:supp_dataset}
\subsection{Dataset Statistics}

\begin{figure*}[t]
    \centering
    \begin{subfigure}[t]{0.49\linewidth}
        \centering
        \includegraphics[width=\linewidth]{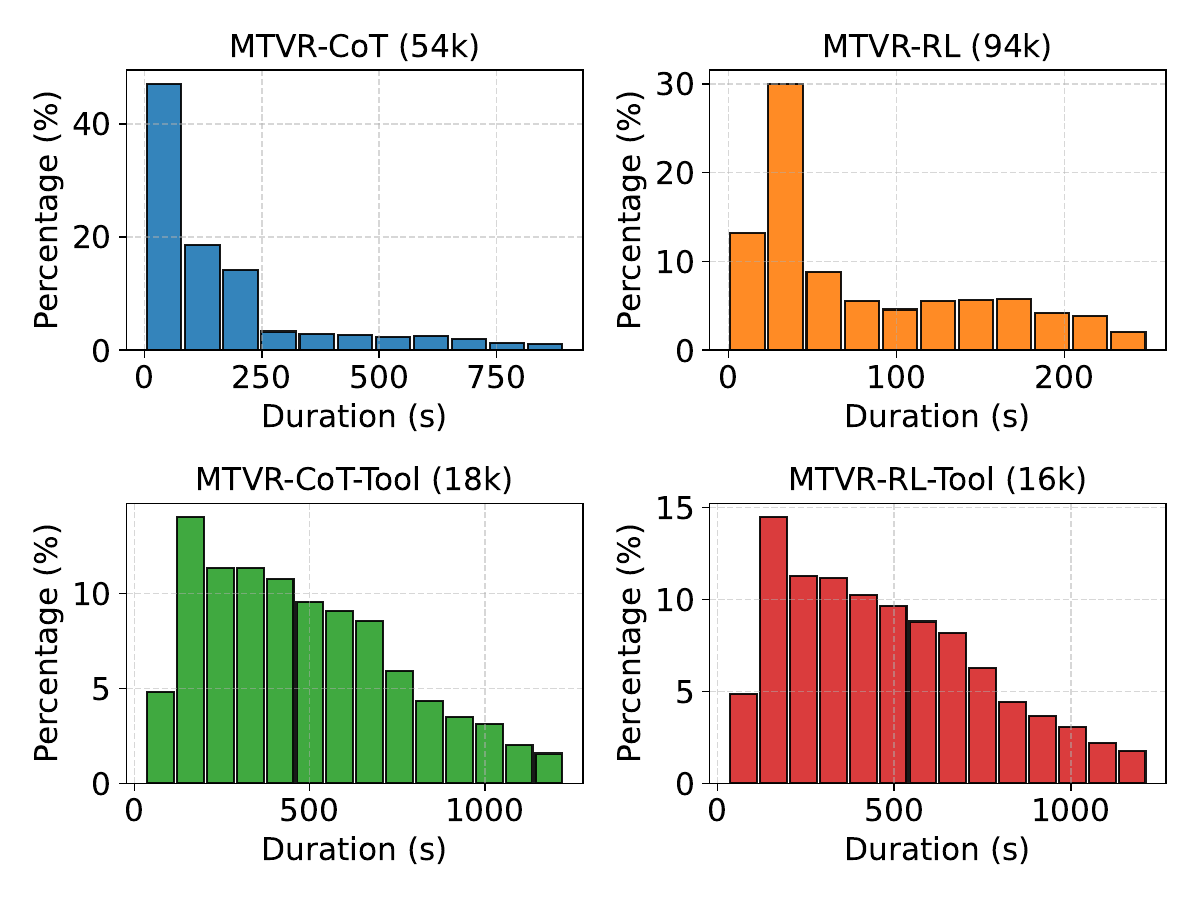}
        \caption{Duration distributions of videos in MTVR dataset.}
        \label{fig:supp_data_duration}
    \end{subfigure}
    \hfill
    \begin{subfigure}[t]{0.49\linewidth}
        \centering
        \includegraphics[width=\linewidth]{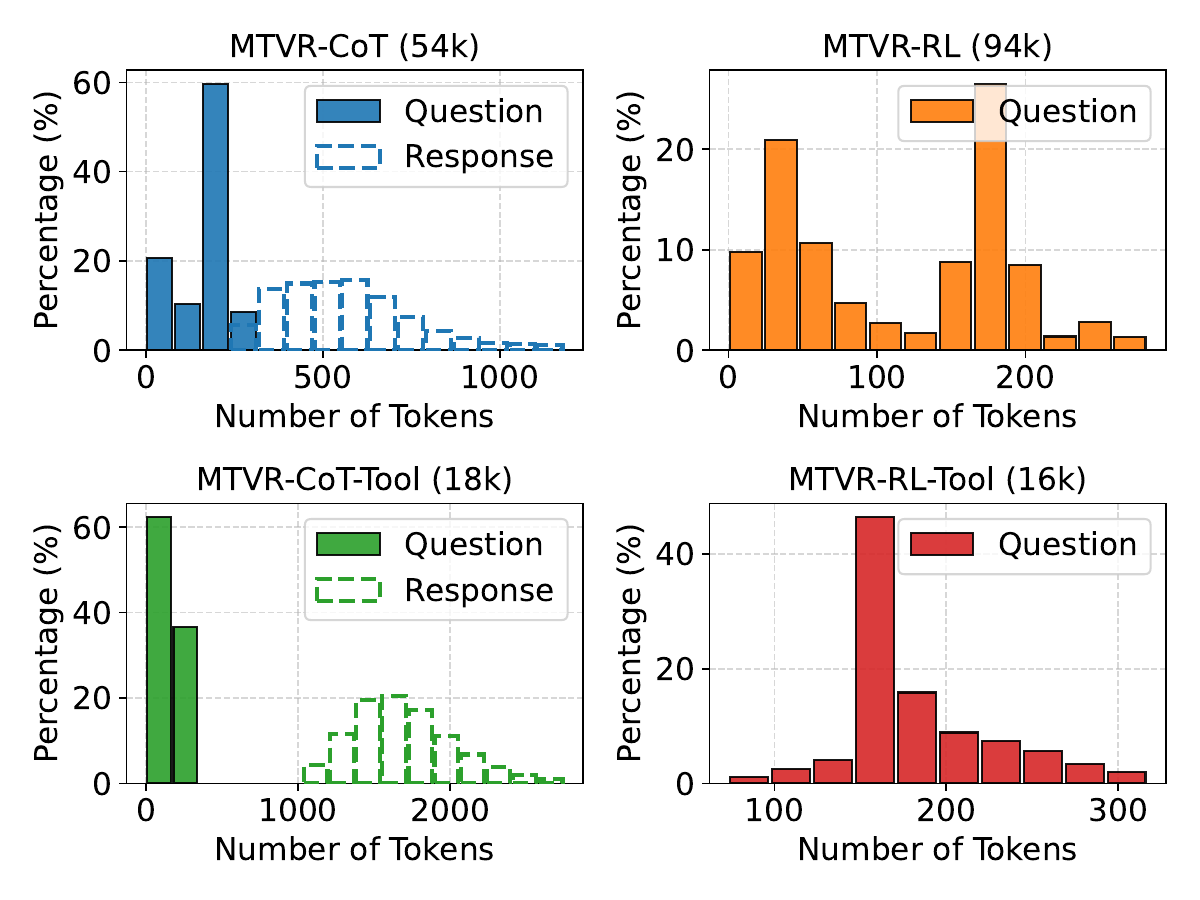}
        \caption{Token length distributions of MTVR dataset.}
        \label{fig:supp_data_token}
    \end{subfigure}
    \caption{Distributions of video duration and token length of the MTVR dataset.}
    \label{fig:supp_data_dist_all}
\end{figure*}

\begin{figure*}[t]
    \centering
    \includegraphics[width=1\linewidth]{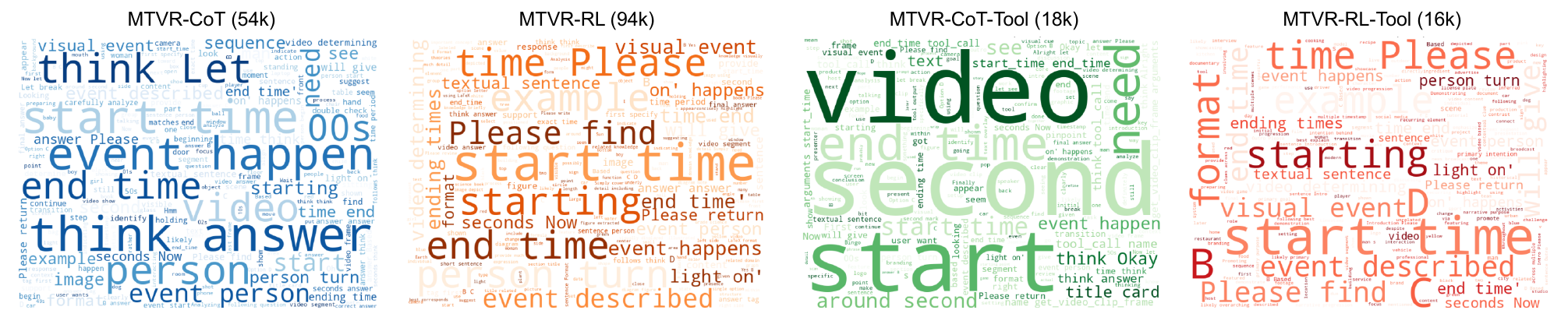}
    \caption{Word cloud of each subset of the MTVR dataset.}
    \label{fig:supp_data_wordcloud}
\end{figure*}

We present comprehensive statistics of the MTVR dataset in \Cref{fig:supp_data_dist_all,fig:supp_data_wordcloud}. 
The dataset comprises four subsets: MTVR-CoT (54k), MTVR-RL (94), MTVR-CoT-Tool (18k), and MTVR-RL-Tool (16k). 
\Cref{fig:supp_data_dist_all} illustrates the distributions of video duration and token length across all subsets. Notably, the video durations vary significantly in each subset, reflecting the diversity of video sources and categories. 
The token length distributions for both questions and responses are also reported, providing insights into the complexity and richness of the textual data. 
Furthermore, \Cref{fig:supp_data_wordcloud} visualizes the most frequent words in each subset using word clouds, highlighting the prevalent concepts and linguistic patterns.

\subsection{Data Filtering Criterion}
\begin{figure*}[t]
    \centering
    \includegraphics[width=1\linewidth]{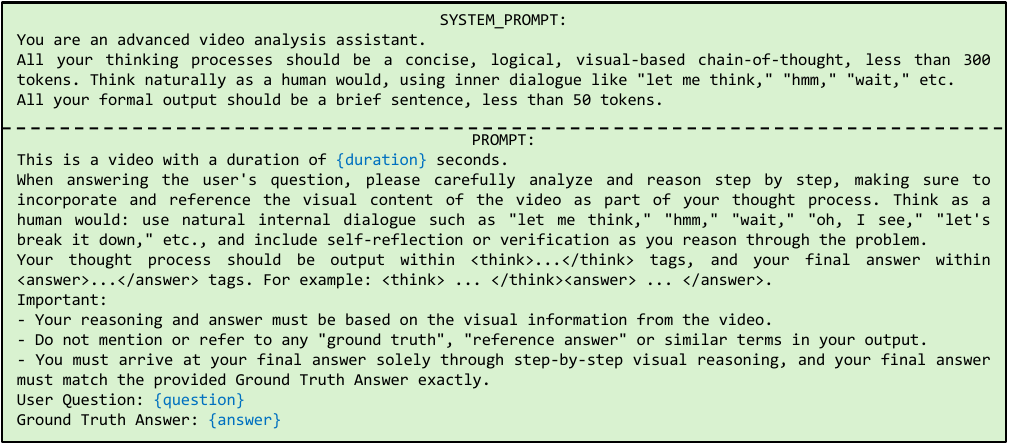}
    \caption{Prompts for text-based CoT generation.}
    \label{fig:supp_prompt1}
\end{figure*}
As described in Sec. 3.2, we propose a rollout filtering process to enhance data quality for DGRPO training. Overly easy or overly hard samples are not beneficial for post-training, as all rollouts for such samples will result in predictions that are either consistently successful or consistently failed. To address this, we design a criterion to filter out these uninformative samples.
Specifically, for a sample $j$ of sub-task $i$, the initial MLLM backbone, i.e., Qwen2.5-VL-7B \cite{bai2025qwen25vl}, generates 8 rollouts. 
The k-th rollout has a task-specific reward $\mathcal{R}_{i,j}^k$ according to \Cref{tab:supp_reward}.
We then compute the reward range for each sample as $\Delta \mathcal{R}_{i,j}=\max \{\mathcal{R}_{i,j}^k\}_{k=1}^{8} - \min \{\mathcal{R}_{i,j}^k\}_{k=1}^{8}$.
Samples with range $\Delta \mathcal{R}_{i,j} \leq 0.05$ are discarded, as they provide limited learning signals for the model during DGRPO process.
The samples that pass the filtering criterion are considered informative and are therefore selected for the MTVR-RL dataset and the next annotation step of the MTVR-CoT dataset.

\subsection{Data Generation Prompts}
\begin{figure*}[!t]
    \centering
    \includegraphics[width=1\linewidth]{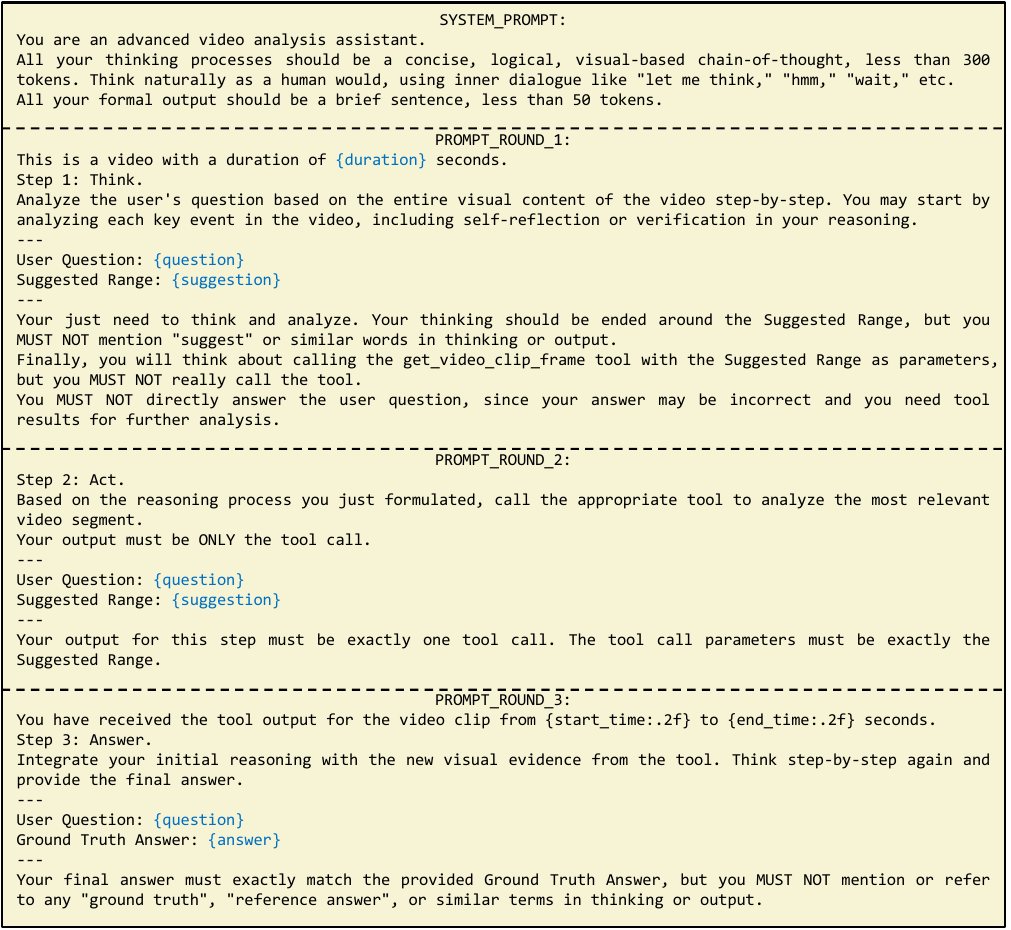}
    \caption{Prompts of each round for multimodal CoT generation.}
    \label{fig:supp_prompt2}
\end{figure*}

During the data generation pipeline, we annotate the text-based CoT reasoning process with the prompts in \Cref{fig:supp_prompt1}, and annotate the multimodal CoT reasoning process with the prompts in \Cref{fig:supp_prompt2} with three round conversations.
In the first round, the reasoning MLLM, e.g., Gemini 2.5 Pro \cite{comanici2025gemini25} is prompted to generate a thinking process. In the second round, it generates the tool call. In the third round, it generates the reflection thinking process and the concluded answer.

Notably, in the second round, the model receives a predefined \textbf{tool parameter suggestion} for the video temporal rounding task, in order to improve the quality of reasoning.
Specifically, for a sample with video $\mathcal{V}$, question $\mathcal{Q}$ and ground truth answer $\mathcal{A}=[s,e]$, the suggested video tool range is calculated as:
\begin{align}
    s^\prime &= \text{clamp}(s - \lambda \cdot |s|\cdot \text{rand}(), 0, L) \\
    e^\prime &= \text{clamp}(e + \lambda \cdot |L-e|\cdot \text{rand}(), 0, L)
\end{align}
Here $L$ denotes the duration of the input video, $\text{rand}()$ is a random float number between 0 and 1. $\lambda$ is a randomness parameter, which is set to 0.2 by default.
The $\text{clamp}$ function ensures that the resulting values remain within the valid range $[0, L]$.
For other tasks like video question answering, we do not use the tool parameter suggestion and remove the corresponding sentences in the prompts in \Cref{fig:supp_prompt2}.
We turn on the thinking mode of reasoning MLLM in round 1 and 3.

\begin{table}[t]
    \centering
    \small
    \begin{tabular}{c|c|ccccc}
        \toprule
        & \multirow{2}{*}{\textbf{Function}} & \multicolumn{1}{c}{\textbf{LVR}} & \multicolumn{1}{c}{\textbf{VidCh}} & \multicolumn{1}{c}{\textbf{MMMU}} & \multicolumn{1}{c}{\textbf{Cha}} & \multirow{2}{*}{\textbf{Avg}} \\
        &  & \textbf{Acc} & \textbf{mIoU} & \textbf{Acc} & \textbf{mIoU} &  \\
        \midrule
        \ding{177} & $\omega_1(D)$ & 70.2 & 28.8 & 52.1 & 57.1 & 52.1 \\
        & $\omega_2(D)$ & 58.3 & 22.6 & 39.9 & 44.2 & 41.3 \\
        & $\omega_3(D)$ & 65.0 & 26.7 & 47.3 & 51.0 & 47.5 \\
        & $\omega_4(D)$ & 68.9 & 27.9 & 49.5 & 54.6 & 50.2 \\
        \bottomrule
    \end{tabular}
    \caption{Ablation study on DGRPO transformations.
    Each experiment contains training for the first two stages, i.e., SFT and DGRPO without tools.
    }
    \label{tab:supp_abl_dgrpo}
\end{table}

\subsection{Data Post-Process}
After generating chain-of-thoughts automatically, we perform a rule-based post-processing to control the data quality.
A sample is excluded from the training dataset if it meets any of the following criteria:
\begin{enumerate}
    \item The chain-of-thought is incomplete or does not reach a final answer.
    \item The generated answer does not match the ground truth.
    \item The sample contains irrelevant or off-topic content, e.g., direct description of ``ground truth'' or ``suggestion''.
\end{enumerate}
After data post-processing, we obtain the final \textbf{MTVR-CoT-72k} dataset, which consists of high-quality and well-formatted samples suitable for cold-start supervised fine-tuning. 
The dataset consists of two subsets, \textbf{MTVR-CoT (54k)} for training stage-1 and \textbf{MTVR-CoT-Tool (18k)} for training stage-3, as illustrated in Fig. 5 in the main paper.

\section{More Ablation Studies}
\label{sec:supp_ablation}
\subsection{Ablation Study on DGRPO Transformations}
We conducted ablation studies on the selection of transformation functions for Sample-wise Difficulty Balance of DGRPO.
As presented in \Cref{sec:supp_method_dgrpo} and \Cref{fig:supp_dgrpo_clamp}, we design four transformation functions:
\begin{align}
    \omega_1(D) &= \text{clamp}(2-D,\,0,\,1)\times 0.5 + 0.5 \\
    \omega_2(D) &= \text{clamp}(2-D,\,0,\,1) \\
    \omega_3(D) &= 1 - 0.25\,D \\
    \omega_4(D) &= 1 - 0.125\,D^2
\end{align}
\Cref{tab:supp_abl_dgrpo} shows the experimental results of using these functions to train for the first two stages.
Comparing $\omega_1(D)$ and $\omega_2(D)$, we observe the importance of soft penalty. $\omega_2(D)$ style hard penalty will totally remove the contribution of easy samples, which is harmful for training.
$\omega_1(D)$ also outperforms continuous functions $\omega_3(D)$ and $\omega_4(D)$. 
We attribute it to the phenomenon that the format reward + tool reward is always equal to 1 after tens of training steps.
After that, the reward will vary from 1.0 to 2.0 in most situations.
Therefore, $\omega_1(D)$ maximizes the range of sample-wise difficulty coefficient, resulting in better performance.

\label{sec:supp_ablation_transform}
\subsection{Ablation Study on Cold Start}
\Cref{tab:supp_abl_coldstart} presents an ablation study on cold start SFT. 
The first two rows serve as baselines without tool usage, where adding cold start SFT shows few performance improvements. 
Introducing additional training with tools (row 3 and row 4) leads to notable improvements. 
Comparison between row 3 and row 4 demonstrates that cold start SFT (row 4) is beneficial for tool-augmented reinforcement learning, 
while directly applying tool-augmented DGRPO after two-stage training (row 3) impedes the model from learning to reason with tool calls.

\begin{table}[ht]
    \centering
    \small
    \setlength{\tabcolsep}{0.7mm}
    \begin{tabular}{l|l|cccccc}
        \toprule
        & \multirow{2}{*}{\textbf{Train Stage}} & \textbf{LVR} & \textbf{VidCh} & \textbf{MMMU} & \textbf{Cha} & \multirow{2}{*}{\textbf{Avg}} & \multirow{2}{*}{\textbf{\#Tools}} \\
        & & \textbf{Acc} & \textbf{mIoU}  & \textbf{Acc}  & \textbf{mIoU} &&\\
        \midrule
        & DGRPO           & 70.1  & 28.5  & 50.9  & 57.5   & 51.8  & 0    \\
        \ding{177} & SFT+DGRPO & 70.2  & 28.8  & 52.1  & 57.1   & 52.1  & 0    \\
        & +DGRPO$^*$      & 70.5  & 30.1  & 51.3  & 57.3   & 52.3  & 0    \\
        \ding{178} & +SFT+DGRPO$^*$  & 79.3  & 35.0  & 54.2  & 59.9   & 57.1  & 0.87 \\
        & SFT+DGRPO$^\dagger$   & 73.7  & 30.9  & 51.3  & 58.9   & 53.7  & 0    \\
        \bottomrule
    \end{tabular}
    \caption{Ablation study on cold start and video resolution.
    The first two rows present results without using tools.
    $^*$ in rows 3 and row 4 indicates further training with tools following row 2 (Exp. \ding{177}).
    $^\dagger$ denotes experiments trained with double the max\_total\_pixels.
    \#Tools is the average number of successful tool calls.
    }
    \label{tab:supp_abl_coldstart}
\end{table}

\subsection{Fair Ablation of Tool Calling}
We conduct an ablation study on whether using tools in Tab. 1 in the main paper (Exp. \ding{177} vs Exp. \ding{178}).
However, this may be due to the different number of video tokens during the final training stage of them, as presented in \Cref{tab:supp_train_config} (Stage-2 and Stage-4).

Therefore, for a fairer comparison, we conduct an additional two-stage experiment, shown in row 5 of \Cref{tab:supp_abl_coldstart}, which matches the experimental setting of row 2 except for the increased video resolution.
In the experiment shown in row 5, we set max\_total\_pixels to 2$\times$64$\times$224$\times$224. This setting ensures that the video token budget is comparable to that of Exp. \ding{178}, as the average number of tool calls in Exp. \ding{178} is less than 1.
As shown in the last two rows of \Cref{tab:supp_abl_coldstart}, VITAL-7B with tools (row 4) still significantly outperforms the improved baseline (row 5) by a large margin.

\subsection{Ablation Study on Data Size}
We provide evaluation results at different steps during the last training stage, DGRPO with tools.
As shown in \Cref{tab:supp_abl_datasize}, although some metric fluctuations occur during training, the overall performance on both long and short video understanding benchmarks consistently improves as the training data size increases.

\begin{table}[ht]
    \centering
    \small
    \setlength{\tabcolsep}{2mm}
    \begin{tabular}{l|l|ccccc}
        \toprule
        & \multirow{2}{*}{\textbf{Data Size}} & \textbf{LVR} & \textbf{VidCh} & \textbf{MMMU} & \textbf{Cha} & \multirow{2}{*}{\textbf{Avg}} \\
        & & \textbf{Acc} &\textbf{ }mIoU & \textbf{Acc} & \textbf{mIoU} &  \\
        \midrule
        & 0   & 68.7 & 18.9 & 45.2 & 52.3 & 46.3 \\
        & 4k  & 73.8 & 27.0 & 50.6 & 55.8 & 51.8 \\
        & 8k  & 76.1 & 31.5 & 51.8 & 58.3 & 54.4 \\
        & 12k & 75.8 & 29.5 & 52.7 & 58.6 & 54.1 \\
        \ding{178} & 16k & 79.3 & 35.0 & 54.2 & 59.9 & 57.1 \\
        \bottomrule
    \end{tabular}
    \caption{Ablation study on different data sizes. Data size denotes the data already used during stage-4 DGRPO training.}
    \label{tab:supp_abl_datasize}
\end{table}

\section{More Case Analyses}
\label{sec:supp_case}
\subsection{Data Quality Analysis}
\begin{figure*}[t]
    \centering
    \includegraphics[width=1\linewidth]{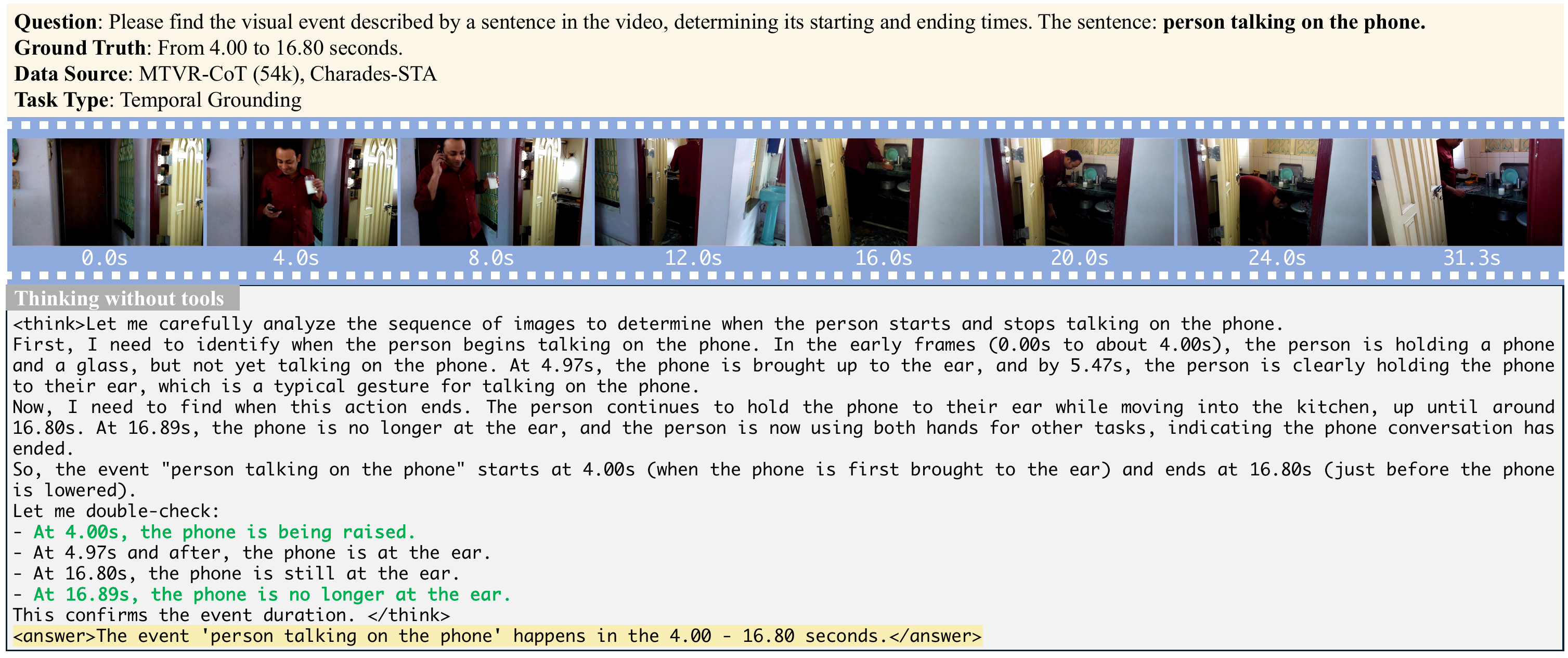}
    \caption{This case study highlights the high data quality of the MTVR-CoT (54k) dataset for temporal grounding tasks. The precise frame-by-frame annotations enable accurate identification of the event boundaries—here, the action "person talking on the phone" is reliably localized between 4.00s and 16.80s. Such detailed and consistent text-based CoT reasoning label supports robust event understanding and temporal modeling in complex video scenarios.}
    \label{fig:supp_case1}
\end{figure*}

\begin{figure*}[t]
    \centering
    \includegraphics[width=1\linewidth]{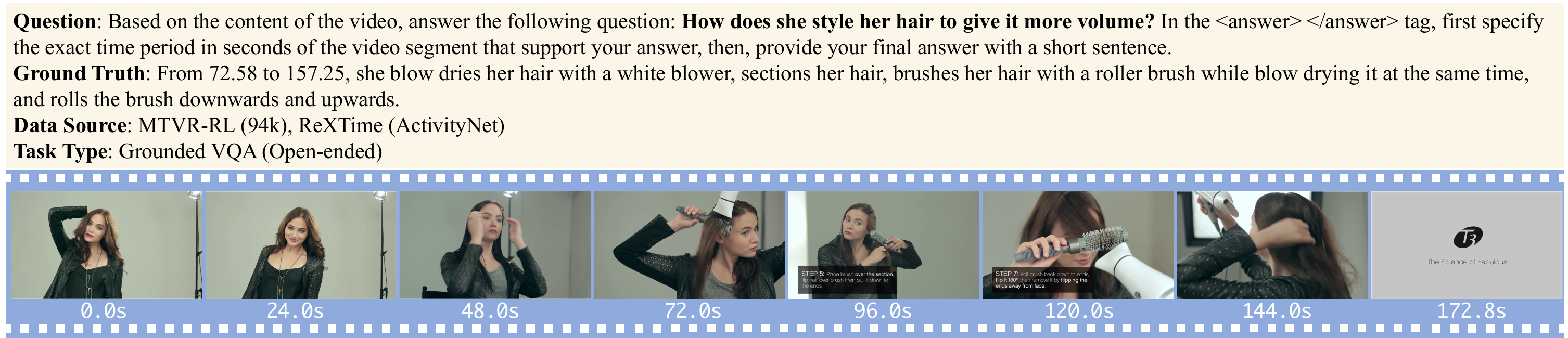}
    \caption{This case study demonstrates the high-quality annotations in the MTVR-RL (94k) dataset for grounded video question answering. The dataset provides precise temporal grounding for open-ended questions, as shown by the accurate localization of the hair styling process between 72.58s and 157.25s. The annotations are collected from original data source.}
    \label{fig:supp_case2}
\end{figure*}

\begin{figure*}[t]
    \centering
    \includegraphics[width=0.98\linewidth]{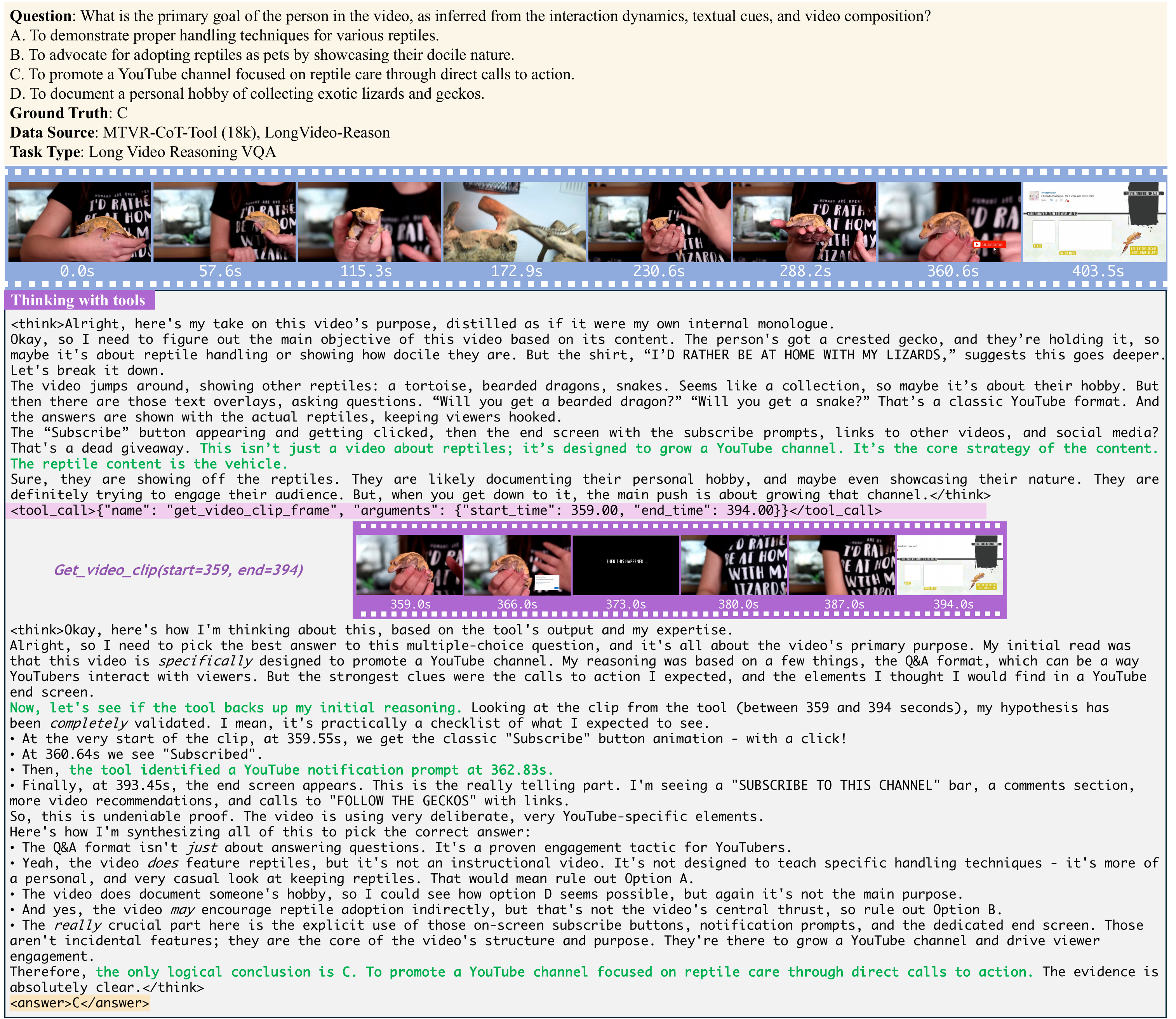}
    \caption{This case study illustrates the depth and reliability of the MTVR-CoT-Tool (18k) dataset for long video reasoning tasks. By combining multimodal cues—including video structure, on-screen text, and explicit calls to action—the dataset enables nuanced analysis of the video's primary goal. The detailed annotation and step-by-step reasoning process allow for accurate identification of the intent: promoting a YouTube channel focused on reptile care through direct viewer engagement. This demonstrates the dataset’s capacity to support complex, high-level, tool-augmented reasoning for long videos.}
    \label{fig:supp_case3}
\end{figure*}

\begin{figure*}[t]
    \centering
    \includegraphics[width=0.98\linewidth]{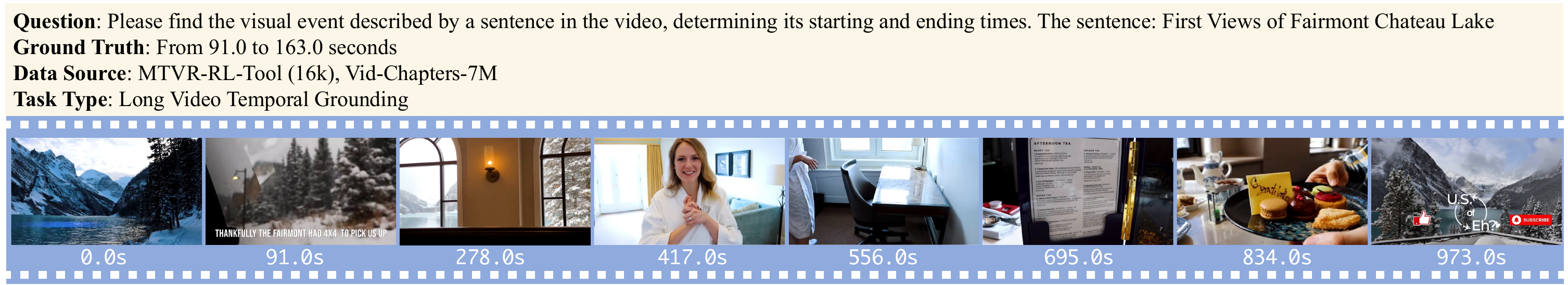}
    \caption{This case study highlights the precise temporal localization enabled by the MTVR-RL-Tool (16k) dataset for long video understanding. By providing accurate annotations for complex events, such as identifying the exact segment (91.0s to 163.0s) for "First Views of Fairmont Chateau Lake," the dataset supports fine-grained reasoning and comprehensive analysis in complex, extended, multi-scene, long video content. The annotations are collected from original data source.}
    \label{fig:supp_case4}
\end{figure*}

High-quality training data is the foundation of robust model performance. In this subsection, we examine representative samples from our MTVR dataset to illustrate the diversity and complexity of the video-question pairs. 
As shown in \Cref{fig:supp_case1,fig:supp_case2,fig:supp_case3,fig:supp_case4}, the MTVR training set features precise temporal annotations, detailed event descriptions, and a wide range of real-world scenarios. 
These examples demonstrate the dataset’s ability to capture fine-grained actions and support complex reasoning, ensuring comprehensive coverage for both short and long video understanding tasks. 
Such high-quality, diverse samples provide a strong foundation for training models capable of robust and generalizable video-language reasoning, e.g., text-based CoT or tool-augmented multimodal CoT.

\begin{figure*}[t]
    \centering
    \includegraphics[width=1\linewidth]{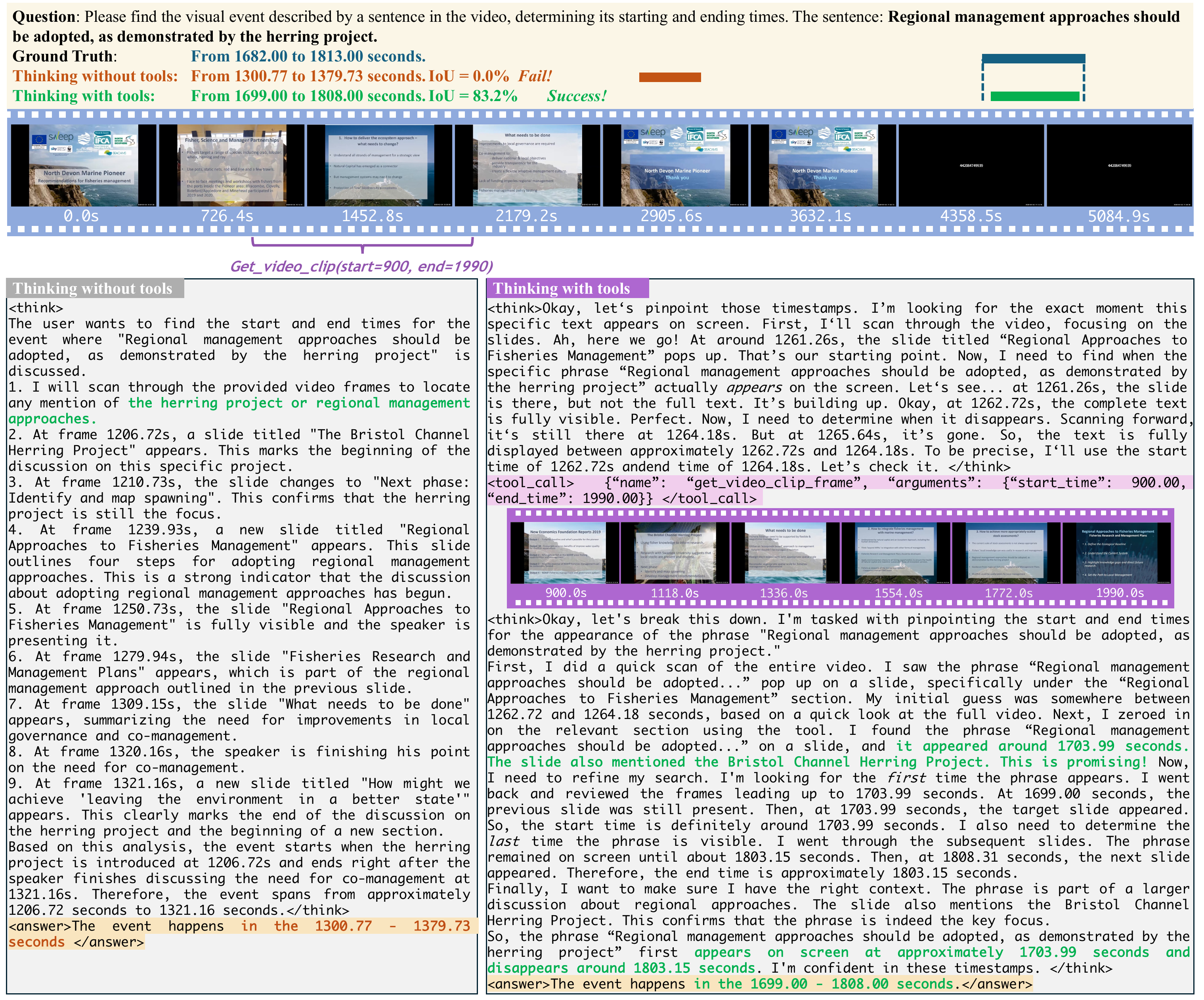}
    \caption{This example demonstrates the effectiveness of the VITAL-7B model's multimodal chain-of-thought (CoT) reasoning for temporal grounding tasks, which leverages tool-augmented analysis to integrate visual and textual information. By systematically using external tools to extract and interpret relevant video segments, the model accurately locates where "regional management approaches should be adopted, as demonstrated by the herring project" is discussed. This tool-augmented multimodal CoT enables robust, step-by-step reasoning and significantly enhances temporal localization in complex, real-world video tasks.}
    \label{fig:supp_case5}
\end{figure*}
\begin{figure*}[t]
    \centering
    \includegraphics[width=1\linewidth]{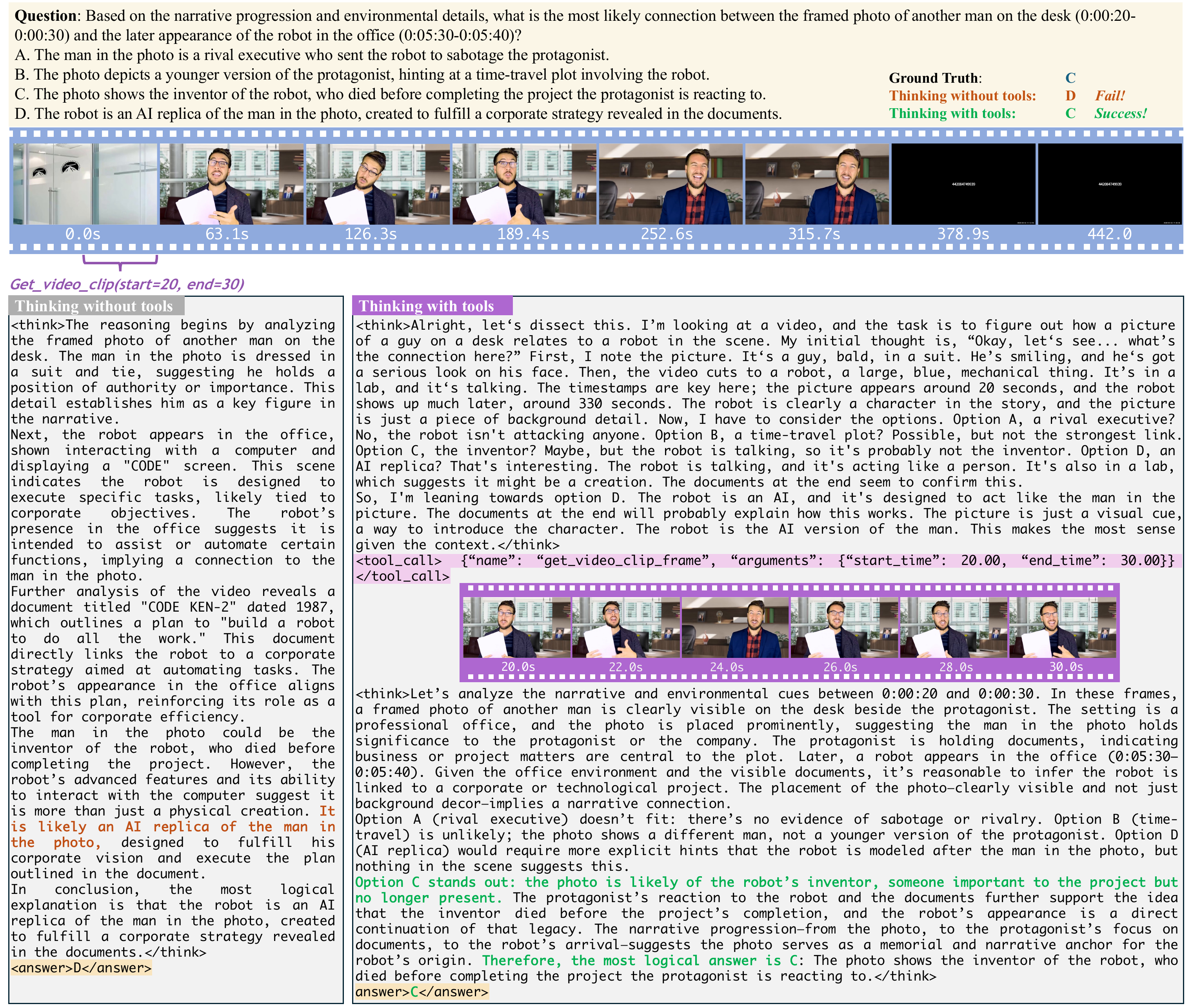}
    \caption{This example highlights the VITAL-7B model's ability to perform tool-augmented multimodal chain-of-thought inference for reasoning VQA tasks. By combining visual scene analysis with textual evidence and leveraging external tools for precise temporal grounding, the model systematically interprets narrative context to infer the relationship between the framed photo and the robot. This approach enables explainable, context-aware decision-making in complex, story-driven video understanding tasks.}
    \label{fig:supp_case6}
\end{figure*}

\subsection{Reasoning Quality Analysis}
Understanding the model's reasoning process is crucial for evaluating its effectiveness in complex video understanding tasks. 
We present several case studies that compare multimodal CoT reasoning chains generated by VITAL-7B with text-based CoT generated by VITAL-7B (w/o). 
As shown in \Cref{fig:supp_case5,fig:supp_case6}, the results demonstrate that tool-augmented multimodal CoT enables VITAL-7B to perform more accurate and fine-grained reasoning compared to text-only CoT. By leveraging both visual and textual modalities, as well as external tools for temporal grounding and evidence extraction, the model can systematically integrate multimodal cues, leading to more precise event localization and deeper narrative understanding. These case studies highlight the substantial advantages of multimodal, tool-augmented reasoning chains for challenging long video video question answering and temporal reasoning tasks.


\section{Limitations and Future Work}
\label{sec:supp_limitations}
\textbf{Limitations.} While VITAL achieves strong results in long video reasoning, our current framework only provides tools for temporal grounding and question answering, limiting its ability to address other tasks such as spatial grounding. Furthermore, our approach mainly focuses on visual features and ignores audio information, which may restrict the model’s overall understanding. Expanding the toolbox and incorporating multimodal features like audio could further enhance the model’s comprehension of videos.

\noindent
\textbf{Future Work.} In the future, we plan to expand the variety of tools within our framework and enable multimodal chain-of-thought reasoning across more modalities, such as integrating both visual and audio cues. This will allow our method to support a broader spectrum of video understanding tasks, including spatio-temporal video grounding, semantic segmentation, and analysis of videos in complex scenarios. Furthermore, we aim to develop more adaptive tool selection strategies to enhance the robustness and versatility of multimodal reasoning in open-ended video understanding.